\documentclass[10pt,letterpaper]{article}
\usepackage[top=0.85in,left=1.75in,footskip=0.75in]{geometry}

\usepackage[sf]{titlesec}

\usepackage{amsmath,amssymb}

\usepackage{changepage}

\usepackage[utf8x]{inputenc}

\usepackage{textcomp,marvosym}

\usepackage{nameref}
\usepackage[colorlinks=true,breaklinks=true]{hyperref}

\usepackage[left, modulo]{lineno}

\usepackage{microtype}
\DisableLigatures[f]{encoding = *, family = * }

\usepackage[table]{xcolor}

\usepackage{array}

\newcolumntype{+}{!{\vrule width 2pt}}

\newlength\savedwidth

\usepackage[aboveskip=1pt,labelfont=bf,labelsep=period,font=small,singlelinecheck=off]{caption}

\makeatletter
\renewcommand{\@biblabel}[1]{\quad#1.}
\makeatother

\usepackage{lastpage,fancyhdr,graphicx}
\usepackage{epstopdf}
\usepackage[ngerman,english]{babel}

\usepackage[round]{natbib}
\usepackage{bibentry}
\nobibliography*

\begin{document}
\vspace*{0.2in}

\begin{flushleft}
{\Large
\textbf\newline{The generalized Hierarchical Gaussian Filter}
}
\newline
\\
Lilian Aline Weber\textsuperscript{1,2,3*},
Peter Thestrup Waade\textsuperscript{4},
Nicolas Legrand\textsuperscript{4},
Anna Hedvig Møller\textsuperscript{4},
Klaas Enno Stephan\textsuperscript{2,5},
Christoph Mathys\textsuperscript{2,6,7}
\\
\bigskip
\textbf{1} Institute for Cognitive Science, Osnabrück University, Osnabrück, Germany
\\
\textbf{2} Translational Neuromodeling Unit, Institute for Biomedical Engineering, University of
Zurich \& ETH Zurich, Zurich, Switzerland
\\
\textbf{3} Department of Psychiatry, University of Oxford, Oxford, United Kingdom
\\
\textbf{4} Cognitive Science Department, Aarhus University, Denmark
\\
\textbf{5} Max Planck Institute for Metabolism Research, Cologne, Germany
\\
\textbf{6} Scuola Internazionale Superiore di Studi Avanzati (SISSA), Trieste, Italy
\\
\textbf{7} Interacting Minds Centre, Aarhus University, Aarhus, Denmark
\\
\bigskip

* lilian.weber@uni-osnabrueck.de

\end{flushleft}

\section*{Abstract}
Hierarchical Bayesian models of perception and learning feature prominently in contemporary cognitive neuroscience where, for example, they inform computational concepts of mental disorders. This includes predictive coding and hierarchical Gaussian filtering (HGF), which differ in the nature of hierarchical representations. In this work, we present a new class of artificial neural networks that unifies computational principles of PC and HGFs. We extend the space of generative models underlying HGF to include a form of nonlinear hierarchical coupling between state values akin to predictive coding and artificial neural networks in general. We derive the update equations corresponding to this generalization of HGF and conceptualize them as connecting a network of (belief) nodes where parent nodes either predict the state of child nodes or their rate of change. This enables us to \emph{(1)} create modular architectures with generic computational steps in each node of the network, and \emph{(2)} disclose the hierarchical message passing implied by generalized HGF models and to compare this to comparable schemes under predictive coding. The practical advances of this work are twofold: on the one hand, our extension allows for a modular construction of ANNs of arbitrarily complex hierarchical structure under the general principles of HGF. On the other hand, by providing a highly flexible implementation of hierarchical Bayesian models available as open source software, it enables new types of empirical data analysis in computational psychiatry.\\

\vfill

\textit{Keywords:} hierarchical Gaussian filter, HGF, predictive coding, artificial neural networks, perceptual inference, neuromodelling, computational psychiatry


\newpage
\section{Introduction}


\subsection{Hierarchical Bayesian models of perception and learning}
Bayesian perspectives on perception have proposed that our brain inverts a hierarchical generative model to infer the causes of its sensory inputs and predict future events \citep{Dayan1995,Rao1999,Friston2010,Doya2011}. Under this ``Bayesian brain'' view, \emph{perception} corresponds to integrating expectations (prior beliefs about hidden states of the world) with incoming sensory information to yield a posterior belief, while \emph{learning} refers to the updating of beliefs about the model's parameters, which takes place more slowly, as experience accumulates. Formally, beliefs are modelled as probability distributions, such that the width of the distribution reflects the uncertainty (inverse precision) associated with that belief. Humans have been shown to take uncertainty into account when combining different sources of information in a manner that conforms to the statistical optimum as prescribed by Bayes' rule \citep{ernst2002,Angelaki2009}.

Because sensory signals are generated by interacting causes in the external environment that span multiple spatial and temporal scales, the brain is assumed to reflect this hierarchy of causes in a correspondingly hierarchical generative model. The Bayesian inversion of this generative model results in a hierarchy of beliefs, where higher levels encode beliefs about increasingly abstract, general, and stable features of the environment. These higher-level beliefs serve as priors for the inference on lower levels. Specifically, at each level of the hierarchy, belief updates serve to reconcile predictions (priors) from higher levels with the actual input (likelihood) from lower levels. Furthermore, under fairly general assumptions \citep[i.e., for all probability distributions from the exponential family,][]{mathys2016,mathys_weber_2020}, these belief updates rest on (precision-weighted) prediction errors (PEs), i.e., the (weighted) mismatch between the model’s predictions and the actual input \citep{Friston2010}.

Popular hierarchical Bayesian models of perception and learning that are built on these ideas are predictive coding \citep{Rao1999,Friston2005} and hierarchical Gaussian filtering \citep{mathys2011,mathys2014}. In these models, estimates of uncertainty are central: the impact of prediction errors on belief updates depends on a precision ratio, which relates the precision of the prior to that of the observation, thus scaling the relative impact that new information has on belief updates. Put simply, mismatches (PEs) elicit stronger belief updates if the prediction about the input (likelihood) is precise, relative to the belief in the current estimate (prior). This form of adaptive scaling, a key element of healthy inference and learning, has also been proposed to lie at the heart of perceptual disturbances observed in mental disorders. For example, an imbalance between the influence of expectations and sensory inputs has featured prominently in attempts to explain the emergence of positive symptoms in schizophrenia, such as hallucinations and delusions \citep{Stephan2006,Fletcher2009,Corlett2009,Corlett2011,Adams2013psycho,Friston2016,sterzer2018,corlett2019hallu}.

The HGF\footnote{we use ``the HGF'' as a shorthand for ``hierarchical Gaussian filtering models of various configurations''} has been particularly useful in this context, as it can be fit to participants' empirically observed behaviour or physiology, and thereby used to infer individual trajectories of precision-weighted PEs and predictions from data. By formulating a response model that links trial-wise perceptual quantities (such as predictions and PEs) to measured quantities (such as choices, reaction times, eye movements, evoked response amplitude in EEG, etc.), the HGF can quantify individual differences in inference and learning in terms of model parameters that encode prior beliefs about higher-order structure in the environment \citep{de_berker_computations_2016,lawson_adults_2017,powers_pavlovian_2017,siegel_computational_2020,sevgi_social_2020,henco_bayesian_2020,rossi-goldthorpe_paranoia_2021,suthaharan_paranoia_2021,kafadar_conditioned_2022,sapey-triomphe_associative_2022,fromm_belief_2023,drusko_novel_2023}. Such a mechanistic characterization of inter-subject variability is of particular interest for fields like Computational Psychiatry, because such differences may explain the heterogeneous nature of psychiatric diseases, and form a basis for mapping them out in a continuous conceptual space or dividing them into more homogeneous subgroups \citep{stephan2014,mathys2016}.

\subsection{Volatility-coupling, value-coupling, and noise-coupling}
The type of belief hierarchy modelled by any particular approach depends on the nature of the generative model. The HGF assumes a particular form of generative model, where hidden states of the world evolve as coupled random walks in time. In current HGF models, the mean (value) of the higher level determines the variance (step size) of the lower level's random walk. In other words, higher levels encode the volatility (or inverse stability) of lower levels (we will call this \textsf{volatility coupling}). This is motivated by the observation that learners must take into account different sources of uncertainty in their belief updates, one of which is the current level of stability in the world: if the world is currently changing (volatile), the agent needs to learn faster. Accordingly, in the HGF, subjective\footnote{Note that the word subjective does not imply conscious accessibility here - it only differentiates the inferred quantities from the ``true`` values as produced by the environment or the generative model.} estimates of increased environmental volatility directly influence the uncertainty associated with lower level beliefs, leading to faster belief updates on the lower level. Previous work has shown that human learners indeed adjust their learning rate according to experimentally manipulated levels of volatility \citep{behrens2007}.

By contrast, predictive coding models typically focus on hierarchies in which higher levels predict the \emph{value} of lower levels. This is implemented as the mean of the probability distribution that represents the lower-level belief, or in other words, the expectation of the value. We will refer to these hierarchies as implementing \textsf{value coupling}. This type of hierarchy is useful for understanding how beliefs about lower-level features depend on higher-level beliefs -- for example, how the perceived brightness of a patch in an image depends on the context (objects, shadows) in which that patch is presented \citep{Rao1999,adelson2005}. Modern predictive coding neural networks are powerful machine learning tools \citep{millidge2022,song2020,whittington2017}. While value coupling - the type of hierarchical coupling in predictive coding - is also often used in theoretical treatments of hierarchical Bayesian modelling, the HGF, in contrast, offers a flexibly applicable implementation of hierarchical beliefs connected by volatility coupling that is being widely used for empirical data analysis.\\

In a noteworthy exception to the typical value hierarchies in predictive coding, \citet{kanai2015} presented a model where higher levels encode the (spatial) precision of lower levels in a static environment. This is different from volatility coupling, where higher levels are concerned with the rate of change on lower levels, but captures a second source of uncertainty in beliefs about hidden states: the level of noise or reliability of the sensory input. Relatedly, in the learning and decision-making literature, two classes of models separately deal with the estimation of process noise \citep[volatility,][]{behrens2007,mathys2011,piray2020a} and observation noise \citep[stochasticity,][]{lee2020,nassar2010}, with recent attempts to capture both sources of uncertainty in a joint model \citep{piray2020b}. Importantly, both volatility and stochasticity estimation can be captured as higher-level belief nodes coupled to the precision of lower-level beliefs.

Here, we show that the HGF can be extended to encompass all of these architectures: precision coupling between hidden states (\textsf{volatility coupling}) and at the observation stage (\textsf{noise coupling}) and, importantly, linear and nonlinear forms of \textsf{value coupling}.  Our extension thus links HGF with predictive coding and also includes a principled way of modelling inference on observation noise. 

The generalised HGF represents a new class of artificial neural networks that unifies computational principles of predictive coding and HGF. Practically, it allows for modular construction of ANNs with complex hierarchical structure, and it provides a flexible tool for empirical data analysis in computational psychiatry via implementation in open software. These theoretical and practical advances will be unpacked in the following sections.

\newpage
\section{Theory: Introducing value coupling}
\label{sec:ghgf-mean}

\subsection*{The generative model for value coupling}
The HGF assumes that an agent is trying to infer on (and learn about) a continuous uncertain quantity $x$ in their environment, which moves (changes) over time. Without any information about the specific form of movement, a generic way of describing movement of a continuous quantity is a Gaussian random walk:
\begin{equation}
    x^{(k)} \sim \mathcal{N}\left(x^{(k-1)}, \, \vartheta\right).
\end{equation}
In the original formulation \citep{mathys2011,mathys2014}, coupling between environmental states at different hierarchical levels was implemented in the form of \textsf{volatility coupling}, where the step size $\vartheta$ (or rate of change/volatility) of a state $x_a$ varies a function of a higher-level state $x_{\check{a}}$:
\begin{equation}
    \vartheta = f\left(x_{\check{a}}^{(k)})\right) = \exp\left(\kappa_{\check{a},a} \, x_{\check{a}}^{(k)} + \omega_a\right),
\end{equation}
with parameters $\kappa_{\check{a},a}$ (scaling the impact of volatility parent $x_{\check{a}}$ on $x_a$) and $\omega_a$ (capturing the "tonic" step size or volatility, which does not vary with time). 
By simultaneously inferring on the state $x_a$ and its rate of change $x_{\check{a}}$, the agent can learn faster (slower) in times when $x_a$ is changing more (less). We call $x_{\check{a}}$ a \textsf{volatility parent} of $x_a$.\\

In contrast, here, we consider the case where a higher-level state $x_b$ influences the value (mean) of the lower-level state $x_a$:
\begin{equation} \label{eq:linear}
	x_a^{(k)} \sim \mathcal{N}\left(x_a^{(k-1)} + f\left(x_b^{(k)}\right), \, \vartheta\right),
\end{equation}
where
\begin{equation} \label{eq:valuecoupling}
    f\left(x_b^{(k)}\right) = \alpha_{b,a} \, g_{b,a}\left(x_b^{(k)}\right) + \rho_a
\end{equation}
is the coupling function with parameter~$\alpha_{b,a}$ (scaling the impact of \textsf{value parent}~$x_b$ on~$x_a$), the constant drift parameter~$\rho_a$, and the function~$g$. The state's mean now evolves as a function of its previous value~$x_a^{(k-1)}$, and a function of state~$x_b^{(k)}$. Crucially, as we will show below, the HGF can deal with both linear and nonlinear transformation functions~$g$, as long as the function~$g$ is twice continuously differentiable almost everywhere. States~$x_b$ and~$x_a$ interact via \textsf{value coupling}.

\begin{figure}
	\includegraphics[width=\textwidth]{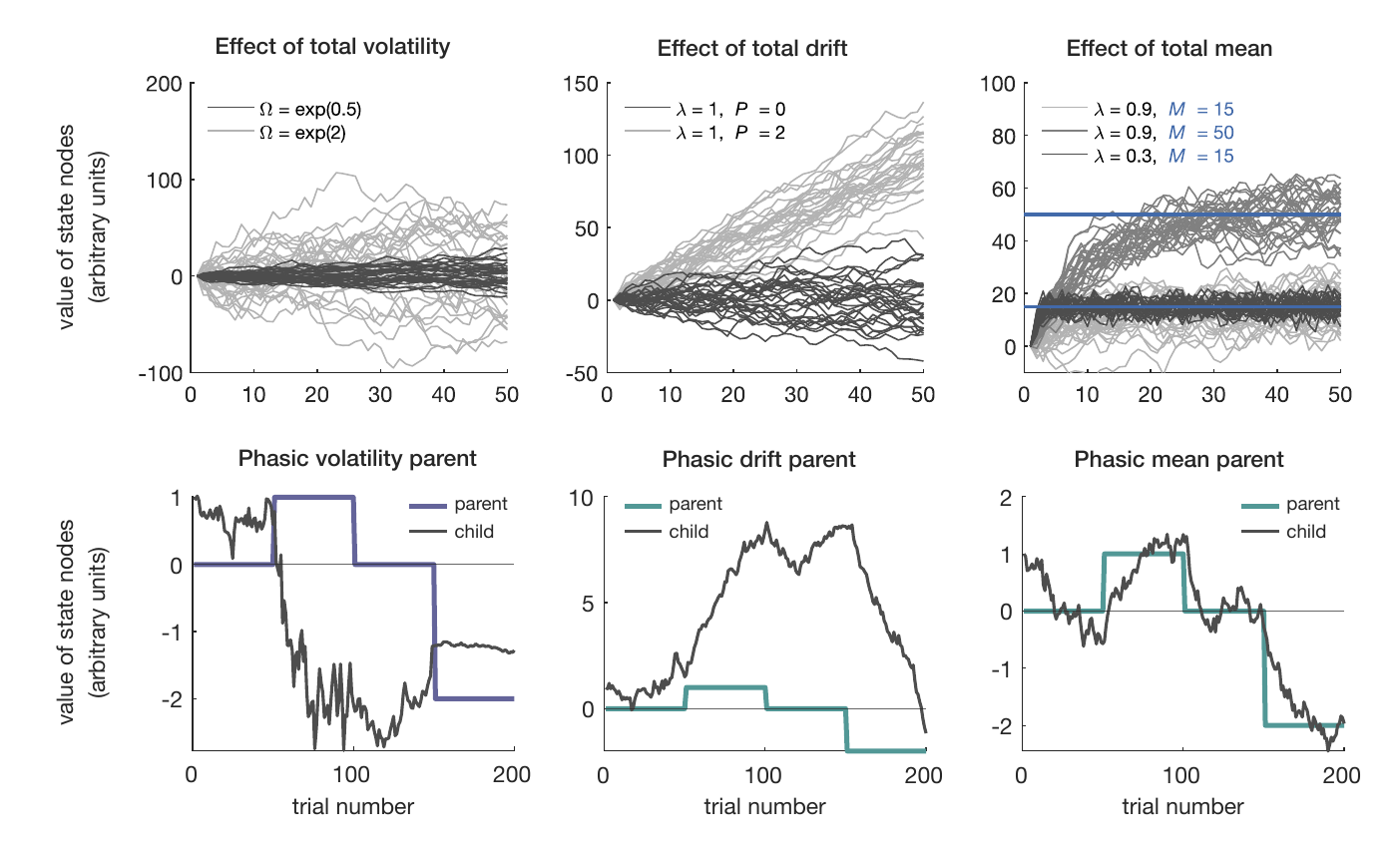}
	\caption{Effects of different coupling types within the generative model of the HGF. \textit{Upper row:} Plots illustrate the effects of a state's total (constant) volatility (left), drift (middle) and mean (right) on the state's evolution over time steps. Each plot shows 30 simulated state trajectories per value of volatility/drift/mean for a continuous state performing a Gaussian random walk over 50 time steps. \textit{Lower row:} Plots illustrate the effects of phasic changes in a state's volatility (left), drift (middle), and mean (right). Each plot shows the trajectory of the parent node and one simulated trajectory of the child node coupled to the parent via volatility/drift/mean coupling.}
\label{fig:randomWalks}
\end{figure}

The effect of a higher-level state on a lower-level state with this kind of coupling can be understood as a phasic drift signal (see Figure~\ref{fig:randomWalks} for illustration). The drift parameter~$\rho$ determines the ``tonic'' drift, equivalent to the tonic volatility parameter~$\omega$. We can use the same configuration also to capture situations where the higher-level state represents a mean value to which the lower-level state reverts back to over time. This is achieved simply by inserting parameter~$\lambda_a \in [0, 1]$ that encodes the state's auto-connection strength:
\begin{equation}
	x_a^{(k)} \sim \mathcal{N}\left(\lambda_a x_a^{(k-1)} + f\left(x_b^{(k)}\right), \, \vartheta\right).
\end{equation} 
When $\lambda_a < 1$, the state $x_a$ will change as an autoregressive process, reverting back to the total mean $M_a$, which is given by:
\begin{equation}
	M_a = \frac{f\left(x_b^{(k)}\right)} {1-\lambda_a},
\end{equation} 
and the value of $\lambda$ will determine the speed with which it does so. In other words, the meaning of $\rho$ and $x_b$ for the evolution of $x_a$ depends on $\lambda$. The difference between the cases $\lambda = 1$ and $\lambda < 1$ is illustrated in Figure~\ref{fig:randomWalks}
. Finally, if $\lambda = 0$, the state does not perform a Gaussian random walk around its own previous mean anymore, instead, its values are normally distributed around a constant or the value of a parent state (reminiscent of an observation of the parent with Gaussian noise).

Importantly, different forms of coupling can be present at the same time: A state $x_a$ can have both a \textsf{volatility parent} $x_{\check{a}}$ (generating changes in its rate of change) and a \textsf{value parent} $x_b$ (generating changes in its mean value). It can also have a drift parameter $\rho_a$ which is a constant influencing its mean -- equivalent to its tonic volatility parameter $\omega_a$ which determines its step size in the absence of a phasic volatility influence. Finally, we allow for inputs to arrive at irregular intervals; therefore, we multiply the total variance of the random walk and the total mean drift by the time~$t^{(k)}$ that has passed since the arrival of the previous sensory input at index $k-1$ \citep{mathys2014}. Together with suitably chosen priors on parameters and initial states \citep[see][]{mathys2014}, the following equation forms the generative model for a state $x_a$ with both volatility and value parent:
\begin{equation} \label{eq:linear2}
	x_a^{(k)} \sim \mathcal{N}\left( \lambda_a x_a^{(k-1)} + t^{(k)} \left(\rho_a + \alpha_{b,a} g_{b,a}\left(x_b^{(k)}\right)\right), \, t^{(k)} \exp\left(\omega_a + \kappa_{\check{a},a} x_{\check{a}}^{(k)}\right)\right).
\end{equation}

In the even more general case, a state could have multiple value parents and multiple volatility parents, each affecting the mean value and rate of change of state $x_a$ in proportion to their respective coupling strengths\footnote{We are here only considering the additive effect of multiple parents on a given state, but more sophisticated interactions are conceivable.}:

\begin{equation} \label{eq:linear3}
	x_a^{(k)} \sim \mathcal{N}\left( \lambda_a x_a^{(k-1)} + \underbrace{t^{(k)} \left(\rho_a + \sum_i \alpha_{b_i,a} g_{b_i,a}\left(x_{b_i}^{(k)}\right)\right)}_{\text{total drift: } P_a^{(k)}}, \, \underbrace{t^{(k)} \exp\left(\omega_a + \sum_j \kappa_{\check{a}_j,a} x_{\check{a}_j}^{(k)}\right)}_{\text{total volatility: } \Omega_a^{(k)}}\right).
\end{equation}

  \graphicspath{{figures/fig1_genmod/}}
  \def\figlabel{fig:genmod}
  \begin{figure}
  \centering

  \small

  \newcommand{\w}[1]{\textcolor{white}{#1}}
  \def\svgwidth{\textwidth}

\begingroup%
  \makeatletter%
  \providecommand\color[2][]{%
    \errmessage{(Inkscape) Color is used for the text in Inkscape, but the package 'color.sty' is not loaded}%
    \renewcommand\color[2][]{}%
  }%
  \providecommand\transparent[1]{%
    \errmessage{(Inkscape) Transparency is used (non-zero) for the text in Inkscape, but the package 'transparent.sty' is not loaded}%
    \renewcommand\transparent[1]{}%
  }%
  \providecommand\rotatebox[2]{#2}%
  \newcommand*\fsize{\dimexpr\f@size pt\relax}%
  \newcommand*\lineheight[1]{\fontsize{\fsize}{#1\fsize}\selectfont}%
  \ifx\svgwidth\undefined%
    \setlength{\unitlength}{905.47749945bp}%
    \ifx\svgscale\undefined%
      \relax%
    \else%
      \setlength{\unitlength}{\unitlength * \real{\svgscale}}%
    \fi%
  \else%
    \setlength{\unitlength}{\svgwidth}%
  \fi%
  \global\let\svgwidth\undefined%
  \global\let\svgscale\undefined%
  \makeatother%
  \begin{picture}(1,0.47042169)%
    \lineheight{1}%
    \setlength\tabcolsep{0pt}%
    \put(0,0){\includegraphics[width=\unitlength]{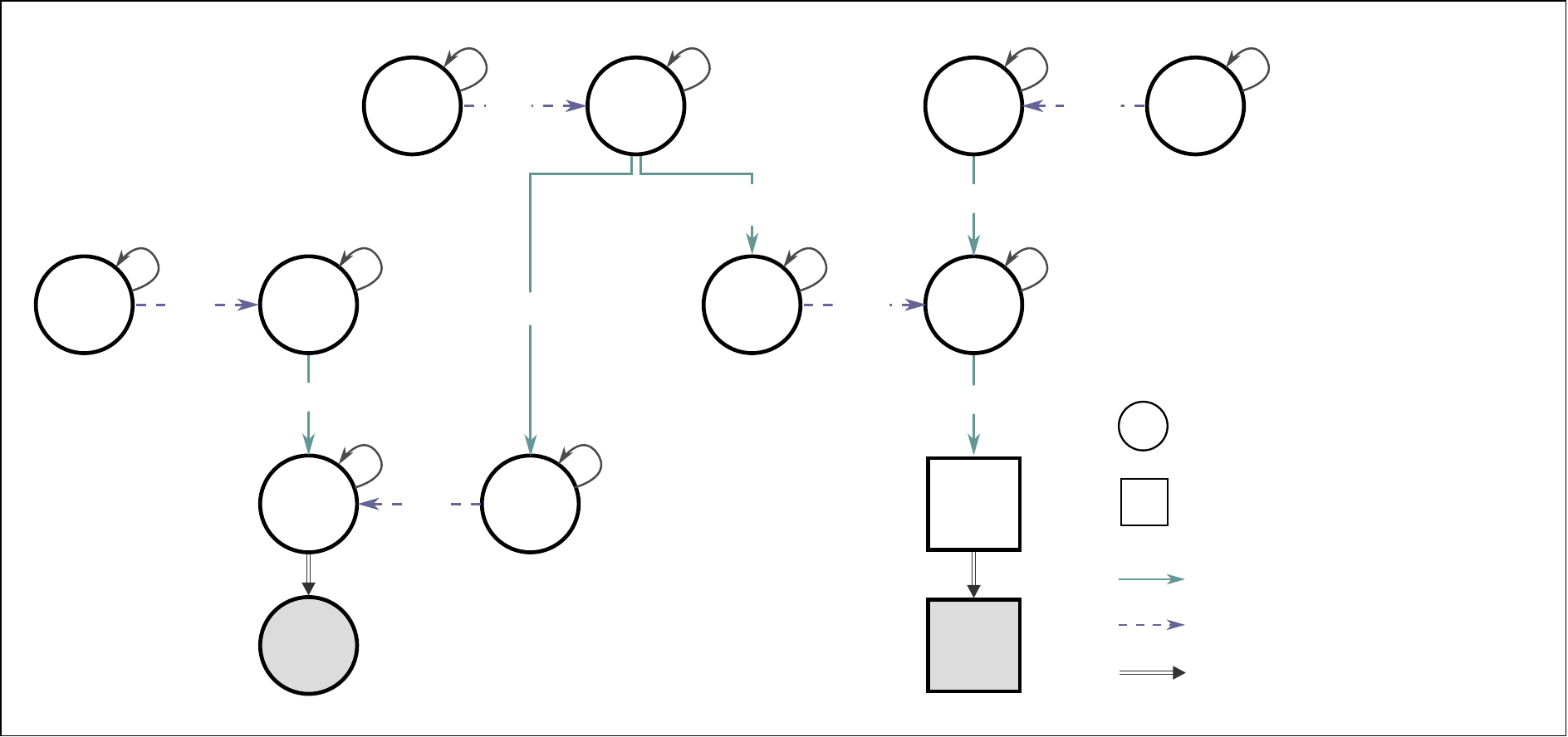}}%
    \put(0.19699938,0.05376604){\color[rgb]{0,0,0}\makebox(0,0)[t]{\lineheight{0}\smash{\begin{tabular}[t]{c}$u_a$\end{tabular}}}}%
    \put(0.19697687,0.14410878){\color[rgb]{0,0,0}\makebox(0,0)[t]{\lineheight{0}\smash{\begin{tabular}[t]{c}$x_a$\end{tabular}}}}%
    \put(0.62158337,0.05376604){\color[rgb]{0,0,0}\makebox(0,0)[t]{\lineheight{0}\smash{\begin{tabular}[t]{c}$u_b$\end{tabular}}}}%
    \put(0.62253429,0.14312723){\color[rgb]{0,0,0}\makebox(0,0)[t]{\lineheight{0}\smash{\begin{tabular}[t]{c}$x_b$\end{tabular}}}}%
    \put(0.05379229,0.27134747){\color[rgb]{0,0,0}\makebox(0,0)[t]{\lineheight{0}\smash{\begin{tabular}[t]{c}$x_{\check{c}}$\end{tabular}}}}%
    \put(0.1979503,0.27134747){\color[rgb]{0,0,0}\makebox(0,0)[t]{\lineheight{0}\smash{\begin{tabular}[t]{c}$x_c$\end{tabular}}}}%
    \put(0.48100632,0.27134747){\color[rgb]{0,0,0}\makebox(0,0)[t]{\lineheight{0}\smash{\begin{tabular}[t]{c}$x_{\check{d}}$\end{tabular}}}}%
    \put(0.62253429,0.27134747){\color[rgb]{0,0,0}\makebox(0,0)[t]{\lineheight{0}\smash{\begin{tabular}[t]{c}$x_d$\end{tabular}}}}%
    \put(0.62158249,0.21236924){\color[rgb]{0.39215686,0.58823529,0.58823529}\rotatebox{-0.00014495}{\makebox(0,0)[t]{\lineheight{0}\smash{\begin{tabular}[t]{c}$\alpha_{d,b}$\end{tabular}}}}}%
    \put(0.26156115,0.39879954){\color[rgb]{0,0,0}\makebox(0,0)[t]{\lineheight{0}\smash{\begin{tabular}[t]{c}$x_{\check{f}}$\end{tabular}}}}%
    \put(0.40593322,0.3973551){\color[rgb]{0,0,0}\makebox(0,0)[t]{\lineheight{0}\smash{\begin{tabular}[t]{c}$x_f$\end{tabular}}}}%
    \put(0.62153834,0.3973551){\color[rgb]{0,0,0}\makebox(0,0)[t]{\lineheight{0}\smash{\begin{tabular}[t]{c}$x_e$\end{tabular}}}}%
    \put(0.62156124,0.33930729){\color[rgb]{0.39215686,0.58823529,0.58823529}\rotatebox{-0.00014495}{\makebox(0,0)[t]{\lineheight{0}\smash{\begin{tabular}[t]{c}$\alpha_{e,d}$\end{tabular}}}}}%
    \put(0.76306626,0.39879954){\color[rgb]{0,0,0}\makebox(0,0)[t]{\lineheight{0}\smash{\begin{tabular}[t]{c}$x_{\check{e}}$\end{tabular}}}}%
    \put(0.33850487,0.14410878){\color[rgb]{0,0,0}\makebox(0,0)[t]{\lineheight{0}\smash{\begin{tabular}[t]{c}$x_{\check{a}}$\end{tabular}}}}%
    \put(0.69727038,0.39857001){\color[rgb]{0.39215686,0.39215686,0.58823529}\rotatebox{-0.00014495}{\makebox(0,0)[t]{\lineheight{0}\smash{\begin{tabular}[t]{c}$\kappa_e$\end{tabular}}}}}%
    \put(0.48008278,0.33930727){\color[rgb]{0.39215686,0.58823529,0.58823529}\rotatebox{-0.00014495}{\makebox(0,0)[t]{\lineheight{0}\smash{\begin{tabular}[t]{c}$\alpha_{f,\check{d}}$\end{tabular}}}}}%
    \put(0.33858921,0.27202258){\color[rgb]{0.39215686,0.58823529,0.58823529}\rotatebox{-0.00014495}{\makebox(0,0)[t]{\lineheight{0}\smash{\begin{tabular}[t]{c}$\alpha_{f,\check{a}}$\end{tabular}}}}}%
    \put(0.19700368,0.21236934){\color[rgb]{0.39215686,0.58823529,0.58823529}\rotatebox{-0.00014495}{\makebox(0,0)[t]{\lineheight{0}\smash{\begin{tabular}[t]{c}$\alpha_{c,a}$\end{tabular}}}}}%
    \put(0.5496378,0.27134749){\color[rgb]{0.39215686,0.39215686,0.58823529}\rotatebox{-0.00014495}{\makebox(0,0)[t]{\lineheight{0}\smash{\begin{tabular}[t]{c}$\kappa_d$\end{tabular}}}}}%
    \put(0.12225974,0.2713476){\color[rgb]{0.39215686,0.39215686,0.58823529}\rotatebox{-0.00014495}{\makebox(0,0)[t]{\lineheight{0}\smash{\begin{tabular}[t]{c}$\kappa_c$\end{tabular}}}}}%
    \put(0.76916014,0.09586841){\color[rgb]{0.39215686,0.58823529,0.58823529}\makebox(0,0)[lt]{\lineheight{0}\smash{\begin{tabular}[t]{l}\textsf{value coupling}\end{tabular}}}}%
    \put(0.76916014,0.06675788){\color[rgb]{0.39215686,0.39215686,0.58823529}\makebox(0,0)[lt]{\lineheight{0}\smash{\begin{tabular}[t]{l}\textsf{volatility coupling}\end{tabular}}}}%
    \put(0.45540578,0.42991264){\color[rgb]{0.29411765,0.29411765,0.29411765}\makebox(0,0)[lt]{\lineheight{0}\smash{\begin{tabular}[t]{l}$\omega_f$\end{tabular}}}}%
    \put(0.31304893,0.42991264){\color[rgb]{0.29411765,0.29411765,0.29411765}\makebox(0,0)[lt]{\lineheight{0}\smash{\begin{tabular}[t]{l}$\omega_{\check{f}}$\end{tabular}}}}%
    \put(0.24603191,0.30221899){\color[rgb]{0.29411765,0.29411765,0.29411765}\makebox(0,0)[lt]{\lineheight{0}\smash{\begin{tabular}[t]{l}$\omega_c$\end{tabular}}}}%
    \put(0.10367507,0.30221899){\color[rgb]{0.29411765,0.29411765,0.29411765}\makebox(0,0)[lt]{\lineheight{0}\smash{\begin{tabular}[t]{l}$\omega_{\check{c}}$\end{tabular}}}}%
    \put(0.24603191,0.17647187){\color[rgb]{0.29411765,0.29411765,0.29411765}\makebox(0,0)[lt]{\lineheight{0}\smash{\begin{tabular}[t]{l}$\omega_a$\end{tabular}}}}%
    \put(0.3862025,0.17647187){\color[rgb]{0.29411765,0.29411765,0.29411765}\makebox(0,0)[lt]{\lineheight{0}\smash{\begin{tabular}[t]{l}$\omega_{\check{a}}$\end{tabular}}}}%
    \put(0.52987325,0.30221899){\color[rgb]{0.29411765,0.29411765,0.29411765}\makebox(0,0)[lt]{\lineheight{0}\smash{\begin{tabular}[t]{l}$\omega_{\check{d}}$\end{tabular}}}}%
    \put(0.67089367,0.30221899){\color[rgb]{0.29411765,0.29411765,0.29411765}\makebox(0,0)[lt]{\lineheight{0}\smash{\begin{tabular}[t]{l}$\omega_d$\end{tabular}}}}%
    \put(0.67089367,0.42991264){\color[rgb]{0.29411765,0.29411765,0.29411765}\makebox(0,0)[lt]{\lineheight{0}\smash{\begin{tabular}[t]{l}$\omega_e$\end{tabular}}}}%
    \put(0.81235472,0.42991264){\color[rgb]{0.29411765,0.29411765,0.29411765}\makebox(0,0)[lt]{\lineheight{0}\smash{\begin{tabular}[t]{l}$\omega_{\check{e}}$\end{tabular}}}}%
    \put(0.27078777,0.14410892){\color[rgb]{0.39215686,0.39215686,0.58823529}\rotatebox{-0.00014495}{\makebox(0,0)[t]{\lineheight{0}\smash{\begin{tabular}[t]{c}$\kappa_a$\end{tabular}}}}}%
    \put(0.32606414,0.39999328){\color[rgb]{0.39215686,0.39215686,0.58823529}\rotatebox{-0.00014495}{\makebox(0,0)[t]{\lineheight{0}\smash{\begin{tabular}[t]{c}$\kappa_f$\end{tabular}}}}}%
    \put(0.76851304,0.14339837){\color[rgb]{0,0,0}\makebox(0,0)[lt]{\lineheight{0}\smash{\begin{tabular}[t]{l}\textsf{binary state}\end{tabular}}}}%
    \put(0.76851304,0.19054028){\color[rgb]{0,0,0}\makebox(0,0)[lt]{\lineheight{0}\smash{\begin{tabular}[t]{l}\textsf{continuous state}\end{tabular}}}}%
    \put(0.76916014,0.03615571){\color[rgb]{0.19607843,0.19607843,0.19607843}\makebox(0,0)[lt]{\lineheight{0}\smash{\begin{tabular}[t]{l}\textsf{generating outcomes}\end{tabular}}}}%
  \end{picture}%
\endgroup%

  \caption{An example of a generative model of sensory inputs with 11 hidden states and two observable outcomes. In this example, the volatility parents $x_{\check{a}}$ and $x_{\check{d}}$ share a value parent $x_f$, which represents a "global" or shared volatility state. Circles -- continuous states, squares -- binary states, observable outcomes -- shaded. Volatility coupling -- dashed lines, value coupling -- straight lines, links of outcomes to their hidden states -- double arrows.}
  \label{\figlabel}
\end{figure}

Because a given state can also be parent to multiple child states at the same time, these extensions allow us to model fairly complex networks of interacting states of the world. 
In Figure~\ref{fig:genmod} we have drawn an example setup with 11 different
environmental states and two outcomes. For this example, and together with priors on parameters and initial states \citep[see][]{mathys2014}, the following equations
describe the generative model (for simplicity, the example uses
linear \textsf{value coupling}, no tonic drifts ($\rho = 0$, $\lambda = 1$), and inputs at regular intervals, i.e.,
$t^{(k)} \equiv 1 \; \forall \, k$ ):
\begin{align}
    u_a^{(k)}           &\sim \mathcal{N}\left(x_a^{(k)}, \varepsilon_u\right)\\
    x_a^{(k)}           &\sim \mathcal{N}\left(x_a^{(k-1)} + \alpha_{c,a} x_c^{(k)}, \exp\left(\kappa_a x_{\check{a}}^{(k)} + \omega_a\right)\right)\\
    x_{\check{a}}^{(k)}   &\sim \mathcal{N}\left(x_{\check{a}}^{(k-1)} + \alpha_{f,\check{a}} x_f^{(k)}, \exp(\omega_{\check{a}})\right)\\
    x_c^{(k)}           &\sim \mathcal{N}\left(x_c^{\left(k-1\right)}, \exp\left(\kappa_c x_{\check{c}}^{(k)} + \omega_c\right)\right)\\
    x_{\check{c}}^{(k)}   &\sim \mathcal{N}\left(x_{\check{c}}^{(k-1)}, \exp\left(\omega_{\check{c}}\right)\right)\\
    u_b^{(k)}           &\sim {\rm Bern} \left(x_b^{(k)}\right)\\
    x_b^{(k)}           &\sim {\rm Bern} \left(S\left(x_d^{(k)}\right)\right)\\
    \end{align}
    \begin{align}
    x_d^{(k)}           &\sim \mathcal{N}\left(x_d^{(k-1)} + \alpha_{e,d} x_e^{(k)},  \exp\left(\kappa_d x_{\check{d}}^{(k)} + \omega_d\right)\right)\\
    x_{\check{d}}^{(k)}   &\sim \mathcal{N}\left(x_{\check{d}}^{(k-1)} + \alpha_{d,f} x_f^{(k)}, \exp(\omega_{\check{d}})\right)\\
    x_e^{(k)}           &\sim \mathcal{N}\left(x_e^{(k-1)},  \exp\left(\kappa_e x_{\check{e}}^{(k)} + \omega_e\right)\right)\\
    x_{\check{e}}^{(k)}   &\sim \mathcal{N}\left(x_{\check{e}}^{(k-1)}, \exp(\omega_{\check{e}})\right)\\
    x_f^{(k)}           &\sim \mathcal{N}\left(x_f^{(k-1)},  \exp\left(\kappa_f x_{\check{f}}^{(k)} + \omega_f\right)\right)\\
    x_{\check{f}}^{(k)}   &\sim \mathcal{N}\left(x_{\check{f}}^{(k-1)}, \exp(\omega_{\check{f}})\right).
\end{align}
Note that this example also includes binary states ($x_b$) and observable outcomes ($u_a$ and $u_b$). Our main discussion will focus on continuous states performing Gaussian random walks, but we will briefly touch on other types of states (binary, categorical, input) in section~\ref{sec:input}.

Using this example network, we introduce two general motifs. First, all states that are value parents of other states (or outcomes) by default have their own volatility parent (and volatility parents therefore share the index with their child node, for example, states~$x_a$ and~$x_{\check{a}}$). Even if in practice, many environmental states might have constant  volatility, from the perspective of the agent, it makes sense to a-priori allow for phasic changes in volatility. From a modelling perspective, these volatility parents could be removed in scenarios with constant volatility.

Second, states that are volatility parents to other states can either have a value parent (as states~$x_{\check{a}}$
and~$x_{\check{d}}$), or no parents (as states~$x_{\check{c}}$,~$x_{\check{e}}$ and~$x_{\check{f}}$). This is because in practice, volatility parents of volatility parents are rarely required. Instead, we suggest that value parents of volatility states
are more useful. In particular, these can be used to model ``global'' or shared volatility states that affect multiple  lower-level volatility beliefs (such as state~$x_f$ in this example, which influences volatility beliefs about  states~$x_a$ and~$x_d$), separately from ``local'' volatility states that only affect speed of change in a single lower-level state (such as the volatility states~$x_{\check{c}}$,~$x_{\check{e}}$ and~$x_{\check{f}}$).\\

In this section, we have introduced value coupling to the generative model of the HGF, for the first time considering the case where the mean of $x_a$ is a function not only of its own previous value but also (some transformation of) the current value of some higher-level state $x_b$, scaled by a coupling parameter $\alpha_{b,a}$. We will now show how an agent can infer on the values of such hidden states.

\subsection*{The belief update equations for value coupling}
An agent employing a generative model of the kind described above to do perceptual inference holds a belief about the current value of each of the states (i.e., every $x$) of this model at every time point $k$.  We describe this belief about state~$x$ at time $k$ as a Gaussian distribution, fully characterized by its mean~$\mu^{(k)}$ and its inverse variance, or precision, $\pi^{(k)}$ at time~$k$.

In the approximate inversion of the generative model for volatility coupling, \citet{mathys2011} derived a set of single-step update equations that represent the approximately Bayes-optimal changes in these beliefs in response to incoming stimuli. Repeating this derivation for the case of value  coupling similarly leads us to simple one-step equations for updating beliefs about states that serve as value parents (for the full derivation of these equations,  cf. \hyperref[sec:appvalue]{Appendix~\ref*{sec:appvalue}}). Assuming state $x_b$ is a value parent to state $x_a$ with a coupling strength $\alpha_{b,a}$, then the new posterior belief about state $x_b$ after observing a new input at time step $k$ is given by:

\begin{equation} \label{eq:ghgfvalueupdate}
\begin{split}
\pi_b^{(k)} &= \hat{\pi}_b^{\left(k\right)} + \hat{\pi}_a^{\left(k\right)} \left(\alpha_{b,a}^2 g_{b,a}'\left(\mu_b^{\left(k-1\right)}\right)^2 - g_{b,a}''\left(\mu_b^{\left(k-1\right)}\right) \delta_a^{\left(k\right)}\right) \\
\mu_b^{\left(k\right)} &= \hat{\mu}_b^{\left(k\right)} + \frac{\hat{\pi}_a^{\left(k\right)} \alpha_{b,a} g_{b,a}'\left(\mu_b^{\left(k-1\right)}\right)}{\pi_b^{\left(k\right)}} \delta_a^{\left(k\right)},
\end{split}
\end{equation}

where $\hat{\pi}_b^{(k)}$ and $\hat{\mu}_b^{(k)}$ refer to the prediction about state $x_b$ before seeing the new input, and $\delta_a^{(k)}$ is the prediction error about the child state $x_a$. 

Note that these equations also hold when introducing the parameter $\lambda$ from above. This parameter will only feature in the computation of the prediction $\hat{\mu}_a^{(k)}$ of state $x_a$ and thus affect the prediction error~$\delta_a^{(k)}$.

In the case of linear value coupling ($g_{b,a}(x) = Ax + B$), the update further simplifies, as $g_{b,a}'(x) = A$ (a factor that we can absorb into $\alpha_{b,a}$) and $g_{b,a}''(x) = 0$:
\begin{equation} \label{eq:ghgfvalueupdatelinear}
\begin{split}
\pi_b^{(k)} &= \hat{\pi}_b^{(k)} + \alpha_{b,a}^2 \hat{\pi}_{a}^{(k)} \\
\mu_b^{(k)} &= \hat{\mu}_b^{(k)} + \frac{\alpha_{b,a} \hat{\pi}_{a}^{(k)}}{\pi_b^{(k)}} \delta_{a}^{(k)}.
\end{split}
\end{equation}

As in the case of volatility coupling, the belief updates are thus driven by precision-weighted prediction errors about the lower belief state in the hierarchy.\\

To get an intuition for these update equations in the case of nonlinear value coupling,
let us consider an example where state~$x_a$ acts as a rectified linear
unit (ReLU), that is, we choose the function~$g$ as the rectifier function, a
popular activation function for deep neural networks
\citep{Hahnloser2000,Lecun2015}:
\begin{equation}
	g_{b,a}(x_b) := \max(0, x_b).
\end{equation}
The first and second derivative of $g$ are then
\begin{equation}
	g_{b,a}'(x_b) = [x_b > 0] := 
	\begin{cases}
    	1, & \text{if } x_b > 0 \\
    	0, & \text{otherwise}
	\end{cases}
\end{equation}
and
\begin{equation}
	g_{b,a}''(x_b) = \delta(x_b=0),
\end{equation}
respectively. For our purposes, we can treat the second derivative as
\begin{equation}
    g_{b,a}''(x_b) \equiv 0.
\end{equation}
Plugging this into equation~\ref{eq:ghgfvalueupdate}, we get
\begin{equation}
	\pi_b^{(k)} = \hat{\pi}_b^{(k)}
            + \alpha_{b,a}^2 \hat{\pi}_{a}^{(k)} \left[\mu_{b}^{(k-1)} > 0\right]
\end{equation}
and
\begin{equation}
    \mu_b^{(k)} = \hat{\mu}_b^{(k)}
			+ \frac{\alpha_{b,a} \hat{\pi}_{a}^{(k)} \left[\mu_{b}^{(k-1)} > 0\right]}
			{\pi_b^{(k)}} \delta_{a}^{(k)}.
\end{equation}
In other words, the impact of lower-level prediction errors $\delta_{a}^{(k)}$ on the posterior belief at the higher level $\mu_b^{(k)}$ and $\pi_b^{(k)}$ depends on the previous state of the higher-level node, such that beliefs only change in response to inputs if they were above zero (or ``active'') in the first place.\\

Similarly, we can construct the reverse coupling function to model saturation:
\begin{equation}
	g_{b,a}(x_b) := \min(0, x_b - \nu)
\end{equation}
Here, state $x_b$ only exerts an influence on the lower-level state~$x_a$, if its own value is below a threshold~$\nu$. In the inference on state~$x_b$, this means that the belief about the parent node,~$mu_b$, will stop being updated after crossing this threshold~$\nu$ (i.e., after becoming saturated):
\begin{equation}
	\pi_b^{(k)} = \hat{\pi}_b^{(k)}
            + \alpha_{b,a}^2 \hat{\pi}_{a}^{(k)} \left[\mu_{b}^{(k-1)} < \nu\right]
\end{equation}
and
\begin{equation}
    \mu_b^{(k)} = \hat{\mu}_b^{(k)}
			+ \frac{\alpha_{b,a} \hat{\pi}_{a}^{(k)} \left[\mu_{b}^{(k-1)} < \nu\right]}
			{\pi_b^{(k)}} \delta_{a}^{(k)}.
\end{equation}

This cessation of updates after crossing a threshold does not mean that a node will remain stuck in place for all future time points. The thresholding only affects the HGF's update step while the prediction step keeps affecting $\hat{\pi}_b^{(k)}$ and $\hat{\mu}_b^{(k)}$, eventually possibly moving back to a region where updates take place again.

These examples demonstrate that our extension of the HGF allows us to build deep networks with non-linear coupling functions useful for both artificial and physiological neural networks.

\newpage
\section{Belief updates in the HGF: A network of nodes}
\label{sec:ghgf-node}

\begin{figure}
	\includegraphics[width = \textwidth]{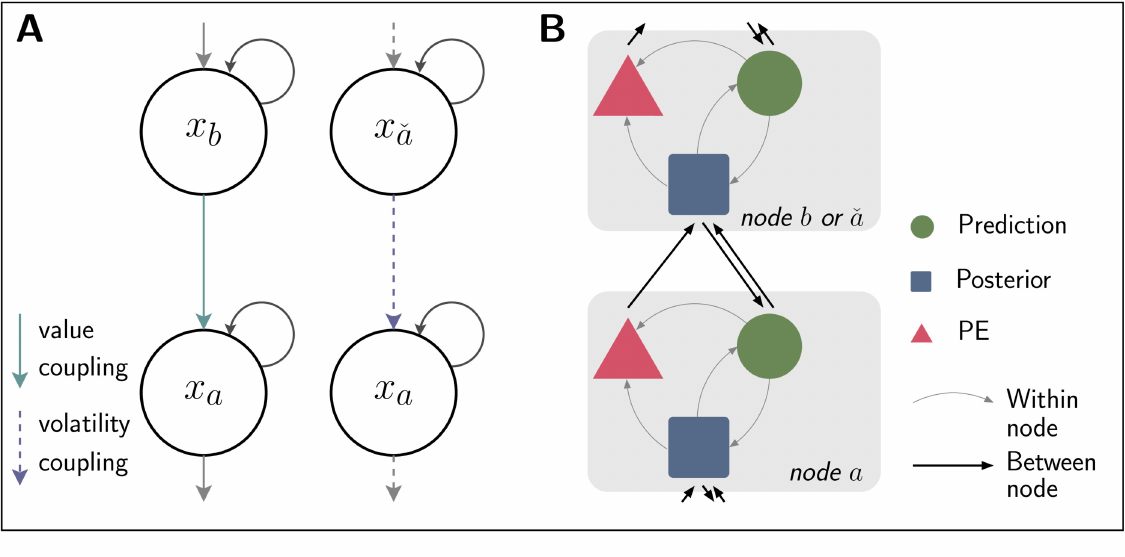}
	\caption{Comparing the flow of information in the generative model of the HGF with the implied belief network. \textbf{\textsf{A}} In the generative model, higher-level states influence the evolution of lower-level states (top-down information flow), either by affecting their mean (value coupling, left) or by changing their evolution rate (volatility coupling, right). \textbf{\textsf{B}} Representation of the message-passing within and between belief nodes as implied by the HGF's belief update equations. New observations cause a cascade of message-passing between nodes that includes bottom-up and top-down information flow. Higher-level beliefs send down their posteriors to inform lower-level predictions. Lower-level belief nodes send prediction errors and the precision of their own prediction bottom-up to drive higher-level belief updating. Within a node, we have placed separate units for the three computational steps that each node has to perform at a given time: the prediction step (green), the update step which results in a new posterior belief (blue), and the prediction error step (red). This message passing scheme generalizes across value and volatility coupling, although the specific messages passed along the connections as well as the computations within the nodes will depend on coupling type (see main text and Figures~\ref{fig:vapeSteps} and~\ref{fig:vopeSteps} for details).}
\label{fig:infnet}
\end{figure}

Just like we can model complex networks of interacting states in the environment using the generative model of the HGF (Figure~\ref{fig:genmod}), we can also think of the inference process of an agent in that environment as a network of interdependent beliefs. The agent entertains a belief about each of the relevant environmental states, and updates these beliefs based on new sensory inputs. Because the agent models its world as a set of hierarchically interacting states, its beliefs about these states will form a hierarchy as well (Figure~\ref{fig:infnet}). Before new inputs arrive, higher-level beliefs inform predictions about lower-level beliefs (top-down information flow), whereas after the arrival of a new piece of information, changes in lower-level beliefs trigger updates of higher-level beliefs (bottom-up information flow). Previous work using the HGF has tended to depict the generative model (Figure~\ref{fig:infnet}\textsf{A}), which then defines the inference model or set of belief update equations. However, the part of the HGF that represents a model of cognition - the evolution of hierarchically interacting beliefs and the relevant flow of information through this hierarchy - is contained in the inference model (or belief update equations, Figure~\ref{fig:infnet}\textsf{B}).

We conceptualize each belief modelled by the HGF as a node in a network, where belief updates involve computations within nodes as well as message passing between nodes. The specific within-node computations and messages passed between nodes will depend on the nature of the coupling. Putting the equations for value and volatility coupling side by side discloses a modular network architecture. This has important consequences for the implementation of the inference model (both in a computer and in a brain), which we summarize in Figure~\ref{fig:infnet}\textsf{B}. We will mainly focus on continuous HGF nodes here - nodes which represent beliefs about continuous quantities that evolve in time as Gaussian random walks. Subsequently, we will consider other types of nodes in HGF models, including categorical nodes and input nodes.\\

We start by noting that the computations of any node within a time step can be subdivided into three steps, an \textsf{update step}, where a new posterior belief is computed based on a prediction and an incoming input or prediction error (PE), a \textsf{PE step}, where the difference between expectation (prediction) and new posterior is computed for further message passing upwards, and a \textsf{prediction step}, where the new posterior is used to predict the value at the next time step. These can be ordered in time as shown in the box:\\

\noindent
\fbox{
  \begin{minipage}{0.7\textwidth}
  	\linespread{2}
  	\textsc{Node} \textit{a} \textsc{at time step} \textit{k}\\
  	\vspace{0.1cm}\\
    \hspace*{0.2cm} (\texttt{re})\texttt{compute} $\mathrm{prediction}^{(k)}_a$ \\
    $\leftarrow$ \texttt{receive} $\mathrm{PE}^{(k)}_{child}$
                from \textit{child} node\\
    
    \textsf{UPDATE step}\\
    \hspace*{0.2cm} \texttt{compute} $\mathrm{posterior}^{(k)}_a$ \\
    \hspace*{0.2cm} \textit{given:} $\mathrm{PE}^{(k)}_{child}$ and 
    								$\mathrm{prediction}^{(k)}_a$
	\vspace{4pt}\\
    $\rightarrow$ \texttt{send} $\mathrm{posterior}^{(k)}_a$
    						 to \textit{child} node
    \vspace{0.2cm}\\
    
    \textsf{PE step} \\
    \hspace*{0.2cm} \texttt{compute} $\mathrm{PE}^{(k)}_a$ \\
    \hspace*{0.2cm} \textit{given:} $\mathrm{prediction}^{(k)}_a$ and 
    								$\mathrm{posterior}^{(k)}_a$
	\vspace{4pt}\\
    $\rightarrow$ \texttt{send} $\mathrm{PE}^{(k)}_a$
    						 to \textit{parent} node\\
    $\leftarrow$ \texttt{receive} $\mathrm{posterior}^{(k)}_{parent}$
    			from \textit{parent} node
    \vspace{0.2cm}\\

    \textsf{PREDICTION step} \\
    \hspace*{0.2cm} \texttt{compute} $\mathrm{prediction}^{(k+1)}_a$ \\
    \hspace*{0.2cm} \textit{given:} $\mathrm{posterior}^{(k)}_a$ and 
    								$\mathrm{posterior}^{(k)}_{parent}$
	\vspace{4pt}\\
  \end{minipage}
}

\vspace{1cm}
\noindent
Two things are worth noting here. First of all, the \textsf{PE step} is a computation that the node performs in service of its parents. From the perspective of the parent node $b$, the $\mathrm{prediction}^{(k)}_a$ represents an expectation of the child node's state, and the $\mathrm{posterior}^{(k)}_a$ corresponds to the actual state of this child at time step $k$. The difference between the two amounts to the prediction error which will serve to update the parent node - in other words, the parent's PE. 

Second, we have placed the \textsf{prediction step} at the end of a time step. This is because usually, we think about the beginning of a time step as starting with receiving a new input, and of a prediction as being present before that input is received. However, in some cases the prediction also depends on the time that has passed in between time steps (e.g., when considering drifts), which is something that can only be evaluated once the new input arrives - hence the additional computation of the (current) prediction at the beginning of the time step. Conceptually, it makes the most sense to think of the prediction as happening continuously between time steps. It is however implementationally more convenient only to compute the prediction once the new input (and with it its arrival time) enters. This ensures both that the posterior means of parent nodes have had enough time to be sent back to their children for preparation for the new input, and that the arrival time of the new input can be taken into account appropriately.\\

A node of the above kind is the first computational sub-unit in our perceptual model, and it can be connected to other nodes via \textsf{volatility} or \textsf{value coupling} depending on the underlying generative model. For node~$a$, another node~$b$ can function as a parent node if the two are connected and node~$b$ represents a belief about a higher-level quantity which affects the belief about node~$a$ according to the generative model.  On the other hand, if node~$b$ refers to a lower-level quantity and is connected to node~$a$, it serves as a child node for node~$a$.

\subsection*{Computations of nodes in the HGF}
In the following, we examine the exact computations at time step $k$ within a node for each of the three steps introduced above. We will compare the relevant computations between \textsf{volatility} and \textsf{value coupling} and identify the messages that have to be sent and received in each step. Since the update step relies on quantities computed in the (previous) prediction step, we start with the computation of the predictions for the current time step (which, as explained above, we think of as being computed prior to the arrival of a new input). We will first only consider the case of linear value coupling (alongside volatility coupling), and then separately examine any differences in these computations for the case of nonlinear value coupling. The results of this analysis are summarized in Figures~\ref{fig:infnet} --~\ref{fig:vopeSteps}.\\

  \graphicspath{{figures/fig3_vape_steps_v2-1/}}
  \def\figlabel{fig:vapeSteps}
  \begin{figure}
  \centering

  \small

  \newcommand{\w}[1]{\textcolor{white}{#1}}
  \def\svgwidth{\textwidth}

\begingroup%
  \makeatletter%
  \providecommand\color[2][]{%
    \errmessage{(Inkscape) Color is used for the text in Inkscape, but the package 'color.sty' is not loaded}%
    \renewcommand\color[2][]{}%
  }%
  \providecommand\transparent[1]{%
    \errmessage{(Inkscape) Transparency is used (non-zero) for the text in Inkscape, but the package 'transparent.sty' is not loaded}%
    \renewcommand\transparent[1]{}%
  }%
  \providecommand\rotatebox[2]{#2}%
  \newcommand*\fsize{\dimexpr\f@size pt\relax}%
  \newcommand*\lineheight[1]{\fontsize{\fsize}{#1\fsize}\selectfont}%
  \ifx\svgwidth\undefined%
    \setlength{\unitlength}{1190.5511811bp}%
    \ifx\svgscale\undefined%
      \relax%
    \else%
      \setlength{\unitlength}{\unitlength * \real{\svgscale}}%
    \fi%
  \else%
    \setlength{\unitlength}{\svgwidth}%
  \fi%
  \global\let\svgwidth\undefined%
  \global\let\svgscale\undefined%
  \makeatother%
  \begin{picture}(1,0.70714286)%
    \lineheight{1}%
    \setlength\tabcolsep{0pt}%
    \put(0,0){\includegraphics[width=\unitlength]{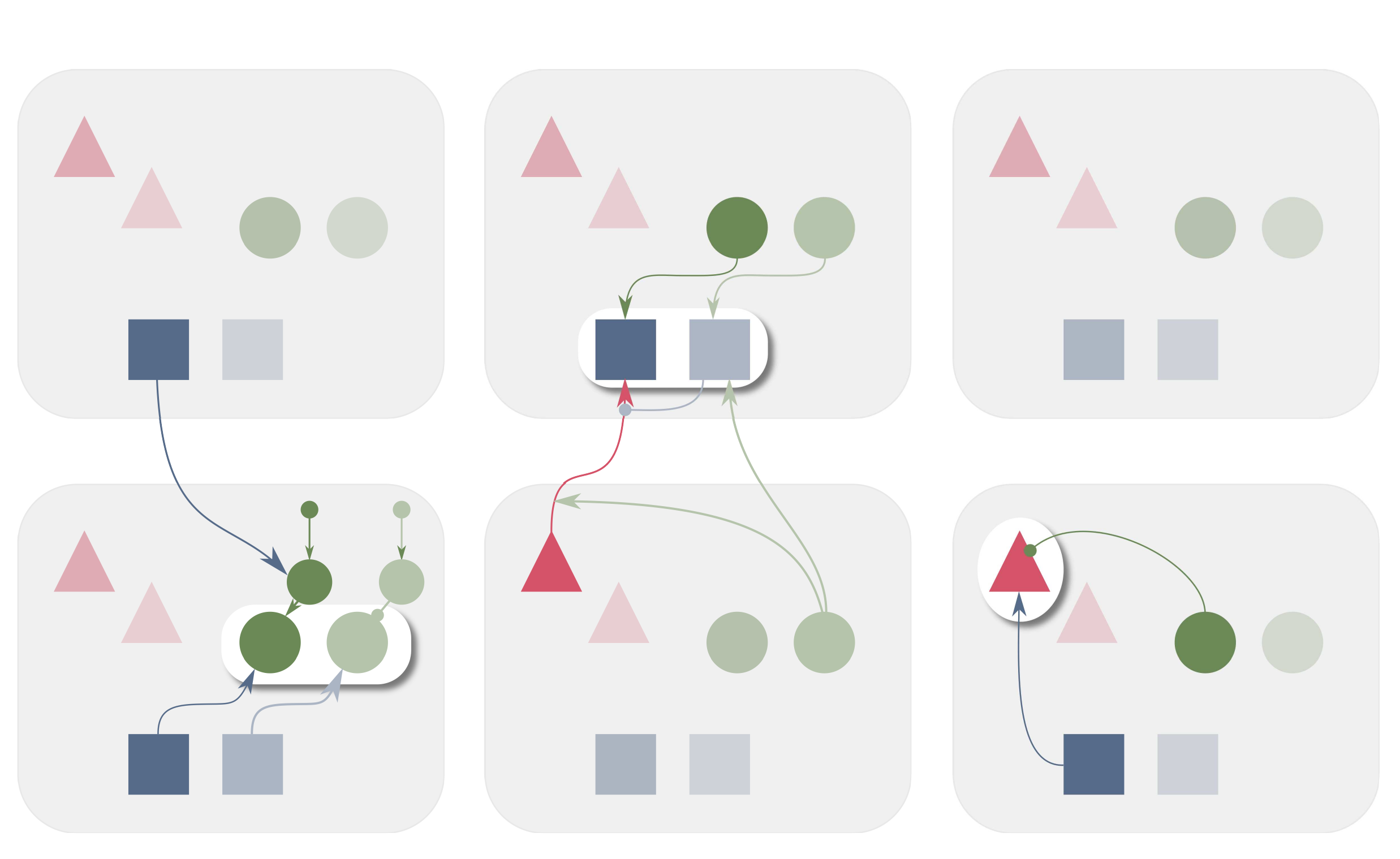}}%
    \put(0.18084479,0.36281164){\color[rgb]{0.58823529,0.58823529,0.58823529}\makebox(0,0)[t]{\lineheight{0}\smash{\begin{tabular}[t]{c}$\pi_{b}$\end{tabular}}}}%
    \put(0.7790419,0.36127972){\color[rgb]{0.58823529,0.58823529,0.58823529}\makebox(0,0)[t]{\lineheight{0}\smash{\begin{tabular}[t]{c}$\mu_{b}$\end{tabular}}}}%
    \put(0.19372469,0.44674967){\color[rgb]{0.58823529,0.58823529,0.58823529}\makebox(0,0)[t]{\lineheight{0}\smash{\begin{tabular}[t]{c}$\hat{\mu}_{b}$\end{tabular}}}}%
    \put(0.25588077,0.44674967){\color[rgb]{0.58823529,0.58823529,0.58823529}\makebox(0,0)[t]{\lineheight{0}\smash{\begin{tabular}[t]{c}$\hat{\pi}_{b}$\end{tabular}}}}%
    \put(0.86012591,0.44800959){\color[rgb]{0.58823529,0.58823529,0.58823529}\makebox(0,0)[t]{\lineheight{0}\smash{\begin{tabular}[t]{c}$\hat{\mu}_{b}$\end{tabular}}}}%
    \put(0.72768742,0.49894361){\color[rgb]{0.58823529,0.58823529,0.58823529}\makebox(0,0)[t]{\lineheight{0}\smash{\begin{tabular}[t]{c}$\delta_{b}$\end{tabular}}}}%
    \put(0.92228199,0.15146979){\color[rgb]{0.58823529,0.58823529,0.58823529}\makebox(0,0)[t]{\lineheight{0}\smash{\begin{tabular}[t]{c}$\hat{\pi}_a$\end{tabular}}}}%
    \put(0.77573567,0.46240591){\color[rgb]{0.58823529,0.58823529,0.58823529}\makebox(0,0)[t]{\lineheight{0}\smash{\begin{tabular}[t]{c}$\Delta_{b}$\end{tabular}}}}%
    \put(0.77573567,0.16712604){\color[rgb]{0.58823529,0.58823529,0.58823529}\makebox(0,0)[t]{\lineheight{0}\smash{\begin{tabular}[t]{c}$\Delta_a$\end{tabular}}}}%
    \put(0.92228199,0.44800959){\color[rgb]{0.58823529,0.58823529,0.58823529}\makebox(0,0)[t]{\lineheight{0}\smash{\begin{tabular}[t]{c}$\hat{\pi}_{b}$\end{tabular}}}}%
    \put(0.84724601,0.36281164){\color[rgb]{0.58823529,0.58823529,0.58823529}\makebox(0,0)[t]{\lineheight{0}\smash{\begin{tabular}[t]{c}$\pi_{b}$\end{tabular}}}}%
    \put(0.84724601,0.06753176){\color[rgb]{0.58823529,0.58823529,0.58823529}\makebox(0,0)[t]{\lineheight{0}\smash{\begin{tabular}[t]{c}$\pi_a$\end{tabular}}}}%
    \put(0.44543009,0.06599984){\color[rgb]{0.58823529,0.58823529,0.58823529}\makebox(0,0)[t]{\lineheight{0}\smash{\begin{tabular}[t]{c}$\mu_a$\end{tabular}}}}%
    \put(0.51363429,0.06753176){\color[rgb]{0.58823529,0.58823529,0.58823529}\makebox(0,0)[t]{\lineheight{0}\smash{\begin{tabular}[t]{c}$\pi_a$\end{tabular}}}}%
    \put(0.52651418,0.15146979){\color[rgb]{0.58823529,0.58823529,0.58823529}\makebox(0,0)[t]{\lineheight{0}\smash{\begin{tabular}[t]{c}$\hat{\mu}_a$\end{tabular}}}}%
    \put(0.44212391,0.16712604){\color[rgb]{0.58823529,0.58823529,0.58823529}\makebox(0,0)[t]{\lineheight{0}\smash{\begin{tabular}[t]{c}$\Delta_a$\end{tabular}}}}%
    \put(0.39407569,0.49894361){\color[rgb]{0.58823529,0.58823529,0.58823529}\makebox(0,0)[t]{\lineheight{0}\smash{\begin{tabular}[t]{c}$\delta_{b}$\end{tabular}}}}%
    \put(0.44212391,0.46240591){\color[rgb]{0.58823529,0.58823529,0.58823529}\makebox(0,0)[t]{\lineheight{0}\smash{\begin{tabular}[t]{c}$\Delta_{b}$\end{tabular}}}}%
    \put(0.06128621,0.49894361){\color[rgb]{0.58823529,0.58823529,0.58823529}\makebox(0,0)[t]{\lineheight{0}\smash{\begin{tabular}[t]{c}$\delta_{b}$\end{tabular}}}}%
    \put(0.10933443,0.46240591){\color[rgb]{0.58823529,0.58823529,0.58823529}\makebox(0,0)[t]{\lineheight{0}\smash{\begin{tabular}[t]{c}$\Delta_{b}$\end{tabular}}}}%
    \put(0.06128621,0.20366374){\color[rgb]{0.58823529,0.58823529,0.58823529}\makebox(0,0)[t]{\lineheight{0}\smash{\begin{tabular}[t]{c}$\delta_a$\end{tabular}}}}%
    \put(0.10933443,0.16712604){\color[rgb]{0.58823529,0.58823529,0.58823529}\makebox(0,0)[t]{\lineheight{0}\smash{\begin{tabular}[t]{c}$\Delta_a$\end{tabular}}}}%
    \put(0.11264063,0.36127972){\color[rgb]{0,0,0}\makebox(0,0)[t]{\lineheight{0}\smash{\begin{tabular}[t]{c}$\mu_{b}$\end{tabular}}}}%
    \put(0.11264063,0.06599984){\color[rgb]{0,0,0}\makebox(0,0)[t]{\lineheight{0}\smash{\begin{tabular}[t]{c}$\mu_a$\end{tabular}}}}%
    \put(0.18084479,0.06753176){\color[rgb]{0,0,0}\makebox(0,0)[t]{\lineheight{0}\smash{\begin{tabular}[t]{c}$\pi_a$\end{tabular}}}}%
    \put(0.14594374,0.11633994){\color[rgb]{0.3372549,0.41960784,0.5372549}\makebox(0,0)[t]{\lineheight{0}\smash{\begin{tabular}[t]{c}$\lambda_a$\end{tabular}}}}%
    \put(0.19372469,0.15146979){\color[rgb]{1,1,1}\makebox(0,0)[t]{\lineheight{0}\smash{\begin{tabular}[t]{c}$\hat{\mu}_a$\end{tabular}}}}%
    \put(0.25588077,0.15146979){\color[rgb]{1,1,1}\makebox(0,0)[t]{\lineheight{0}\smash{\begin{tabular}[t]{c}$\hat{\pi}_a$\end{tabular}}}}%
    \put(0.29058935,0.0343342){\color[rgb]{0,0,0}\makebox(0,0)[rt]{\lineheight{0}\smash{\begin{tabular}[t]{r}node $a$\end{tabular}}}}%
    \put(0.29058935,0.32596061){\color[rgb]{0,0,0}\makebox(0,0)[rt]{\lineheight{0}\smash{\begin{tabular}[t]{r}node $b$\end{tabular}}}}%
    \put(0.44543009,0.36127972){\color[rgb]{1,1,1}\makebox(0,0)[t]{\lineheight{0}\smash{\begin{tabular}[t]{c}$\mu_{b}$\end{tabular}}}}%
    \put(0.51363429,0.36281164){\color[rgb]{1,1,1}\makebox(0,0)[t]{\lineheight{0}\smash{\begin{tabular}[t]{c}$\pi_{b}$\end{tabular}}}}%
    \put(0.52651418,0.44800959){\color[rgb]{0,0,0}\makebox(0,0)[t]{\lineheight{0}\smash{\begin{tabular}[t]{c}$\hat{\mu}_{b}$\end{tabular}}}}%
    \put(0.58867027,0.44800959){\color[rgb]{0,0,0}\makebox(0,0)[t]{\lineheight{0}\smash{\begin{tabular}[t]{c}$\hat{\pi}_{b}$\end{tabular}}}}%
    \put(0.39407569,0.20366374){\color[rgb]{0,0,0}\makebox(0,0)[t]{\lineheight{0}\smash{\begin{tabular}[t]{c}$\delta_a$\end{tabular}}}}%
    \put(0.58867027,0.15146979){\color[rgb]{0,0,0}\makebox(0,0)[t]{\lineheight{0}\smash{\begin{tabular}[t]{c}$\hat{\pi}_a$\end{tabular}}}}%
    \put(0.62337883,0.0343342){\color[rgb]{0,0,0}\makebox(0,0)[rt]{\lineheight{0}\smash{\begin{tabular}[t]{r}node $a$\end{tabular}}}}%
    \put(0.62337883,0.32596061){\color[rgb]{0,0,0}\makebox(0,0)[rt]{\lineheight{0}\smash{\begin{tabular}[t]{r}node $b$\end{tabular}}}}%
    \put(0.43385092,0.29021723){\color[rgb]{0.83529412,0.3254902,0.41176471}\makebox(0,0)[rt]{\lineheight{0}\smash{\begin{tabular}[t]{r}$\alpha_{b,a}$\end{tabular}}}}%
    \put(0.53425511,0.29021723){\color[rgb]{0.70980392,0.76862745,0.67058824}\makebox(0,0)[lt]{\lineheight{0}\smash{\begin{tabular}[t]{l}$\alpha_{b,a}^2$\end{tabular}}}}%
    \put(0.7790419,0.06599984){\color[rgb]{0,0,0}\makebox(0,0)[t]{\lineheight{0}\smash{\begin{tabular}[t]{c}$\mu_a$\end{tabular}}}}%
    \put(0.72768742,0.20366374){\color[rgb]{1,1,1}\makebox(0,0)[t]{\lineheight{0}\smash{\begin{tabular}[t]{c}$\delta_a$\end{tabular}}}}%
    \put(0.86012591,0.15146979){\color[rgb]{0,0,0}\makebox(0,0)[t]{\lineheight{0}\smash{\begin{tabular}[t]{c}$\hat{\mu}_a$\end{tabular}}}}%
    \put(0.95699059,0.0343342){\color[rgb]{0,0,0}\makebox(0,0)[rt]{\lineheight{0}\smash{\begin{tabular}[t]{r}node $a$\end{tabular}}}}%
    \put(0.95699059,0.32596061){\color[rgb]{0,0,0}\makebox(0,0)[rt]{\lineheight{0}\smash{\begin{tabular}[t]{r}node $b$\end{tabular}}}}%
    \put(0.03790344,0.57951888){\color[rgb]{0,0,0}\makebox(0,0)[lt]{\lineheight{0}\smash{\begin{tabular}[t]{l}\textsf{Prediction step}\end{tabular}}}}%
    \put(0.37069293,0.57951888){\color[rgb]{0,0,0}\makebox(0,0)[lt]{\lineheight{0}\smash{\begin{tabular}[t]{l}\textsf{Update step}\end{tabular}}}}%
    \put(0.70430465,0.57951888){\color[rgb]{0,0,0}\makebox(0,0)[lt]{\lineheight{0}\smash{\begin{tabular}[t]{l}\textsf{Prediction error step}\end{tabular}}}}%
    \put(0.118748,0.29021723){\color[rgb]{0.3372549,0.41960784,0.5372549}\makebox(0,0)[rt]{\lineheight{0}\smash{\begin{tabular}[t]{r}$\alpha_{b,a}$\end{tabular}}}}%
    \put(0.23824027,0.22817221){\color[rgb]{0.41960784,0.5372549,0.3372549}\makebox(0,0)[t]{\lineheight{0}\smash{\begin{tabular}[t]{c}$\rho_{a}$\end{tabular}}}}%
    \put(0.2212809,0.19320068){\color[rgb]{0,0,0}\makebox(0,0)[t]{\lineheight{0}\smash{\begin{tabular}[t]{c}$P_{a}$\end{tabular}}}}%
    \put(0.3061116,0.22826121){\color[rgb]{0.70980392,0.76862745,0.67058824}\makebox(0,0)[t]{\lineheight{0}\smash{\begin{tabular}[t]{c}$\omega_{a}$\end{tabular}}}}%
    \put(0.28692549,0.19325595){\color[rgb]{0,0,0}\makebox(0,0)[t]{\lineheight{0}\smash{\begin{tabular}[t]{c}$\Omega_{a}$\end{tabular}}}}%
  \end{picture}%
\endgroup%

  \caption{Message-passing for value coupling. Interactions of two nodes, node~$a$ and its value parent node~$b$, are shown during the three steps of a trial (Prediction step, left; Update step, middle; Prediction error step, right). The quantities that are being computed in each step are highlighted in white. Note that each step, we only show the computations for either the parent or the child node. Connections with arrowheads indicate positive (excitatory) influences, connections with circular heads indicate negative (inhibitory) influences. Arrows ending on units indicate additive influences, those ending on other arrows indicate multiplicative influences. Each HGF quantity that changes across trials is assigned its own unit. Parameters ($\alpha$, $\kappa$, $\lambda$, $\omega$ and $\rho$) determine connection strengths. For clarity, the volatility and drift nodes $\Omega$ and $P$ are only shown during the prediction step.}
  \label{\figlabel}
\end{figure}

\subsubsection*{The prediction step}
In the \textsf{prediction step}, node~$a$ prepares for receiving the new input.
This entails computing a new prediction, based on the previously updated posterior beliefs.
In general, the prediction of a new mean (or the new mean of the predictive
distribution) will depend on whether node~$a$ has any value parents, whereas the
precision of the new prediction will be influenced by the presence and posterior
mean of the node's volatility parents.

In the general case, the mean and precision of the new prediction are computed
as follows:
\vspace{0.5cm}

\noindent
\begin{minipage}{\textwidth}

\begin{align}\label{eq:predict}
	\hat{\mu}_a^{(k)}     &= \lambda_a \mu_a^{(k-1)} + P_a^{(k)}\\
	\hat{\pi}_a^{(k)}     &= \frac{1}{\frac{1}{\pi_a^{(k-1)}} + \Omega_a^{(k)}},
\end{align}
\vspace{1pt}

\end{minipage}
\vspace{0.5cm}

\noindent
where $P_a^{(k)}$ is the \textsf{total predicted drift of the mean} and $\Omega_a^{(k)}$ is
the \textsf{total predicted volatility} (or step size). In the case of value coupling, the relative contribution of the node's own previous value and the predicted drift is determined by the parameter~$\lambda$ (see Section~\ref{sec:ghgf-mean}). Both the total predicted drift as well as the volatility are computed as a sum of a constant (or tonic) term, given by a model parameter, and a time-varying (or phasic) term which is driven by their parents: 

The total predicted mean drift equals the sum of constant term $\rho_a$ (the tonic drift parameter of node~$a$) and the sum of  posterior means of all value parents $\mu_{b_i}^{(k-1)}$ (where parents are indexed by $i$) at $k-1$, weighted by their connection
strengths~$\alpha_{b_i,a}$:
\begin{equation}\label{eq:exdrift}
	P_a^{(k)} = t^{(k)} \left(\rho_a + \sum_{i=1}^{N_{vapa}} \alpha_{b_i,a}
	\mu_{b_i}^{(k-1)}\right),
\end{equation}
where $t^{(k)}$ denotes the time that has passed between~$k-1$ and~$k$, and $N_{vapa}$ is the number of value parents. Similarly, the total predicted volatility~$\Omega_a^{(k)}$ is a function of a constant term~$\omega_a$ (the tonic volatility parameter) and the posterior means of all volatility parents $\mu_{\check{a}_j}^{(k-1)}$ at the previous time point $k-1$, weighted by their connection strengths~$\kappa_{\check{a}_j,a}$:
\begin{equation}\label{eq:exvol}
	\Omega_a^{(k)} = t^{(k)} \exp\left(\omega_a + \sum_{j=1}^{N_{vopa}} \kappa_{\check{a}_j,a}
	\mu_{\check{a}_j}^{(k-1)}\right)
\end{equation}
If node~$a$ does not have any parents, both the predicted drift~$P_a$ and the predicted volatility~$\Omega_a$ are fully determined by constant parameters ($\rho_a$ and $\omega_a$) and the time between subsequent observations. In
 \textsc{hgf}s without drift ($\rho_a = 0$), the predicted mean for the next time step is equal to the posterior mean of the current time step. Equations~\ref{eq:predict} to~\ref{eq:exvol} nicely reflect the roles that value parents and volatility parents play in the generative model, where value parents model a phasic influence on a child node's mean, and volatility parents model a phasic influence on a child node's step size or volatility.\\

In sum, the \textsf{prediction step} for node~$a$ only depends on knowing its own
posterior belief from the previous time step and having received its parents' posteriors in time before the new input arrives. The implied message passing  for this computational step is visualized in the left panel of Figure~\ref{fig:vapeSteps} for value coupling, and Figure~\ref{fig:vopeSteps} for volatility coupling.


\subsubsection*{The update step}
The \textsf{update step} consists of computing a new posterior belief, i.e., a new mean~$\mu^{(k)}$ and a new precision~$\pi^{(k)}$, given a new input from the level (node) below (usually, a prediction error $\delta$), and the node's own prediction ($\hat{\mu}^{(k)}$ and $\hat{\pi}^{(k)}$). In this case, the exact computations within a node depend on the nature of its children: If node~$b$ is the \textsf{value parent} of node~$a$, then the following update equations apply to node~$b$:

\vspace{0.5cm}

\noindent
\begin{minipage}{\textwidth}

\begin{align}\label{eq:linvalueupdate}
\pi_b^{(k)} &= \hat{\pi}_b^{(k)} 
            + \alpha_{b,a}^2 \hat{\pi}_{a}^{(k)}\\
\mu_b^{(k)} &= \hat{\mu}_b^{(k)} 
            + \frac{\alpha_{b,a} \hat{\pi}_{a}^{(k)}} {\pi_b^{(k)}} \delta_{a}^{(k)} \label{eq:updateVapeWithPrecision}
\end{align}         
\vspace{1pt}

\end{minipage}
\vspace{0.5cm}

\noindent
Thus, at the time of the update, node~$i$ needs to have access to the following quantities:
\begin{description}
\item[Its own prediction:]     $\hat{\mu}_b^{(k)}$, $\hat{\pi}_b^{(k)}$
\item[Coupling strength:]       $\alpha_{b,a}$
\item[From level below:]        $\delta_{a}^{(k)}$, $\hat{\pi}_{a}^{(k)}$
\end{description}
All of these are available at the time of the update. Node~$b$ therefore only needs to receive the PE and the precision of the prediction from the child nodes to perform its update. The middle panel of Figure~\ref{fig:vapeSteps} illustrates these computations.

Note that from the equations above, we can define another quantity, the precision-weighted prediction error (pwPE):
\begin{equation}
    \psi_{a,b}^{(k)} = \frac{\hat{\pi}_{a}^{(k)}} {\pi_b^{(k)}} \delta_{a}^{(k)}.
\end{equation}
This quantity summarizes the size of the belief update in node~b due to changes in node~a (before accounting for connection strength) and is often of interest in experimental investigations of neural correlates of prediction errors in belief updating \citep{iglesias_hierarchical_2013, diaconescu_hierarchical_2020,weber2020,weber2022}.

  \graphicspath{{figures/fig4_vope_steps_v2-1/}}
  \def\figlabel{fig:vopeSteps}
  \begin{figure}
  \centering

  \small

  \newcommand{\w}[1]{\textcolor{white}{#1}}
  \def\svgwidth{\textwidth}

\begingroup%
  \makeatletter%
  \providecommand\color[2][]{%
    \errmessage{(Inkscape) Color is used for the text in Inkscape, but the package 'color.sty' is not loaded}%
    \renewcommand\color[2][]{}%
  }%
  \providecommand\transparent[1]{%
    \errmessage{(Inkscape) Transparency is used (non-zero) for the text in Inkscape, but the package 'transparent.sty' is not loaded}%
    \renewcommand\transparent[1]{}%
  }%
  \providecommand\rotatebox[2]{#2}%
  \newcommand*\fsize{\dimexpr\f@size pt\relax}%
  \newcommand*\lineheight[1]{\fontsize{\fsize}{#1\fsize}\selectfont}%
  \ifx\svgwidth\undefined%
    \setlength{\unitlength}{1190.5511811bp}%
    \ifx\svgscale\undefined%
      \relax%
    \else%
      \setlength{\unitlength}{\unitlength * \real{\svgscale}}%
    \fi%
  \else%
    \setlength{\unitlength}{\svgwidth}%
  \fi%
  \global\let\svgwidth\undefined%
  \global\let\svgscale\undefined%
  \makeatother%
  \begin{picture}(1,0.70714286)%
    \lineheight{1}%
    \setlength\tabcolsep{0pt}%
    \put(0,0){\includegraphics[width=\unitlength]{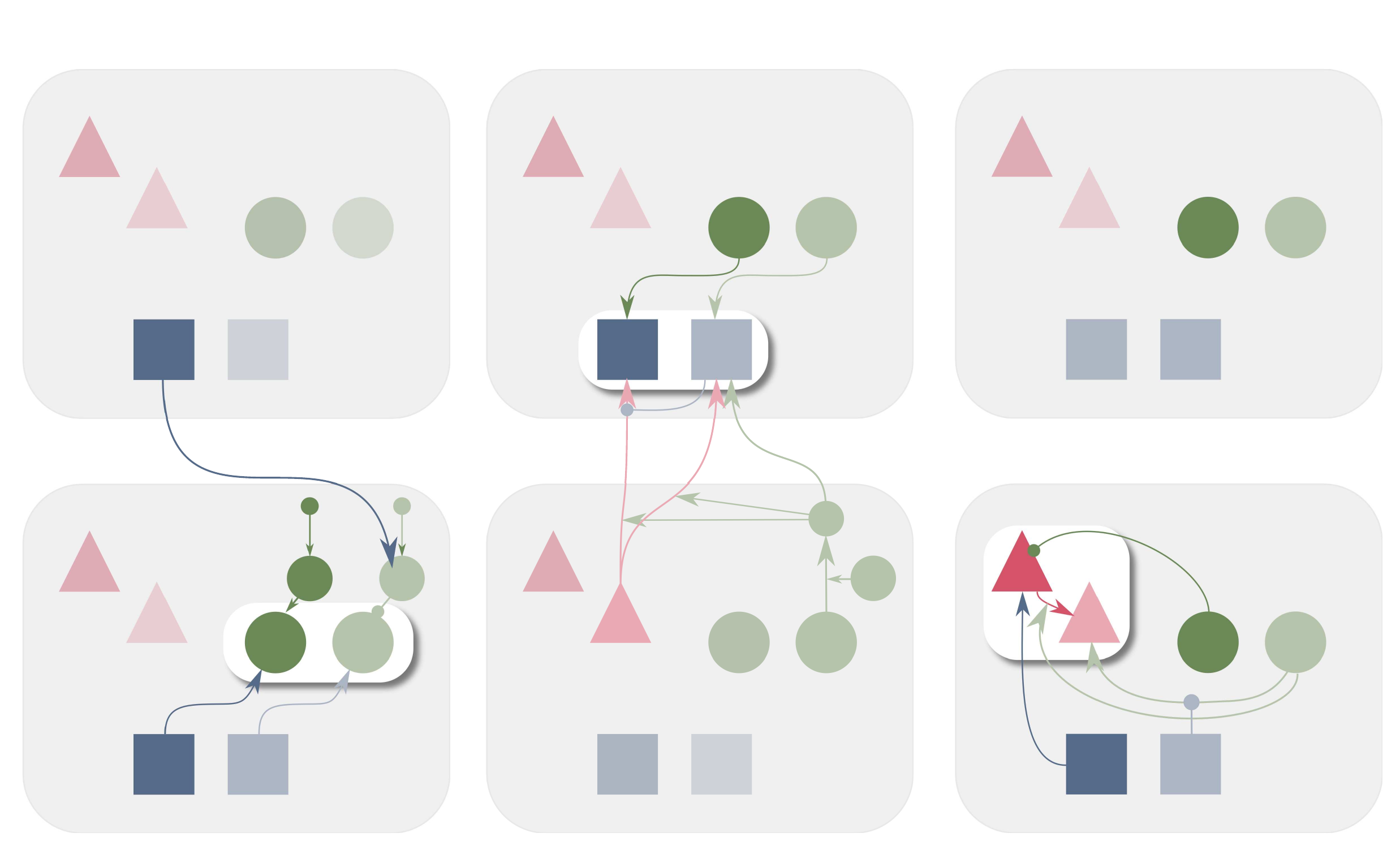}}%
    \put(0.18462455,0.3602918){\color[rgb]{0.58823529,0.58823529,0.58823529}\makebox(0,0)[t]{\lineheight{0}\smash{\begin{tabular}[t]{c}$\pi_{\check{a}}$\end{tabular}}}}%
    \put(0.52792335,0.15405661){\color[rgb]{0.58823529,0.58823529,0.58823529}\makebox(0,0)[t]{\lineheight{0}\smash{\begin{tabular}[t]{c}$\hat{\mu}_a$\end{tabular}}}}%
    \put(0.19750445,0.44926951){\color[rgb]{0.58823529,0.58823529,0.58823529}\makebox(0,0)[t]{\lineheight{0}\smash{\begin{tabular}[t]{c}$\hat{\mu}_{\check{a}}$\end{tabular}}}}%
    \put(0.25966054,0.44926951){\color[rgb]{0.58823529,0.58823529,0.58823529}\makebox(0,0)[t]{\lineheight{0}\smash{\begin{tabular}[t]{c}$\hat{\pi}_{\check{a}}$\end{tabular}}}}%
    \put(0.78086173,0.3614466){\color[rgb]{0.58823529,0.58823529,0.58823529}\makebox(0,0)[t]{\lineheight{0}\smash{\begin{tabular}[t]{c}$\mu_{\check{a}}$\end{tabular}}}}%
    \put(0.73202709,0.49911083){\color[rgb]{0.58823529,0.58823529,0.58823529}\makebox(0,0)[t]{\lineheight{0}\smash{\begin{tabular}[t]{c}$\delta_{\check{a}}$\end{tabular}}}}%
    \put(0.06758581,0.4989436){\color[rgb]{0.58823529,0.58823529,0.58823529}\makebox(0,0)[t]{\lineheight{0}\smash{\begin{tabular}[t]{c}$\delta_{\check{a}}$\end{tabular}}}}%
    \put(0.11311419,0.4624059){\color[rgb]{0.58823529,0.58823529,0.58823529}\makebox(0,0)[t]{\lineheight{0}\smash{\begin{tabular}[t]{c}$\Delta_{\check{a}}$\end{tabular}}}}%
    \put(0.06758581,0.20366373){\color[rgb]{0.58823529,0.58823529,0.58823529}\makebox(0,0)[t]{\lineheight{0}\smash{\begin{tabular}[t]{c}$\delta_a$\end{tabular}}}}%
    \put(0.11311419,0.16712603){\color[rgb]{0.58823529,0.58823529,0.58823529}\makebox(0,0)[t]{\lineheight{0}\smash{\begin{tabular}[t]{c}$\Delta_a$\end{tabular}}}}%
    \put(0.3980047,0.49901058){\color[rgb]{0.58823529,0.58823529,0.58823529}\makebox(0,0)[t]{\lineheight{0}\smash{\begin{tabular}[t]{c}$\delta_{\check{a}}$\end{tabular}}}}%
    \put(0.44353307,0.46247288){\color[rgb]{0.58823529,0.58823529,0.58823529}\makebox(0,0)[t]{\lineheight{0}\smash{\begin{tabular}[t]{c}$\Delta_{\check{a}}$\end{tabular}}}}%
    \put(0.44683926,0.06606682){\color[rgb]{0.58823529,0.58823529,0.58823529}\makebox(0,0)[t]{\lineheight{0}\smash{\begin{tabular}[t]{c}$\mu_a$\end{tabular}}}}%
    \put(0.51504345,0.0650789){\color[rgb]{0.58823529,0.58823529,0.58823529}\makebox(0,0)[t]{\lineheight{0}\smash{\begin{tabular}[t]{c}$\pi_a$\end{tabular}}}}%
    \put(0.3980047,0.20373071){\color[rgb]{0.58823529,0.58823529,0.58823529}\makebox(0,0)[t]{\lineheight{0}\smash{\begin{tabular}[t]{c}$\delta_a$\end{tabular}}}}%
    \put(0.7775555,0.46257313){\color[rgb]{0.58823529,0.58823529,0.58823529}\makebox(0,0)[t]{\lineheight{0}\smash{\begin{tabular}[t]{c}$\Delta_{\check{a}}$\end{tabular}}}}%
    \put(0.11642039,0.36127972){\color[rgb]{0,0,0}\makebox(0,0)[t]{\lineheight{0}\smash{\begin{tabular}[t]{c}$\mu_{\check{a}}$\end{tabular}}}}%
    \put(0.11642039,0.06599985){\color[rgb]{0,0,0}\makebox(0,0)[t]{\lineheight{0}\smash{\begin{tabular}[t]{c}$\mu_a$\end{tabular}}}}%
    \put(0.18462455,0.06501193){\color[rgb]{0,0,0}\makebox(0,0)[t]{\lineheight{0}\smash{\begin{tabular}[t]{c}$\pi_a$\end{tabular}}}}%
    \put(0.19750445,0.15398964){\color[rgb]{1,1,1}\makebox(0,0)[t]{\lineheight{0}\smash{\begin{tabular}[t]{c}$\hat{\mu}_a$\end{tabular}}}}%
    \put(0.25966054,0.15398964){\color[rgb]{1,1,1}\makebox(0,0)[t]{\lineheight{0}\smash{\begin{tabular}[t]{c}$\hat{\pi}_a$\end{tabular}}}}%
    \put(0.29436911,0.03433421){\color[rgb]{0,0,0}\makebox(0,0)[rt]{\lineheight{0}\smash{\begin{tabular}[t]{r}node $a$\end{tabular}}}}%
    \put(0.29436911,0.3259606){\color[rgb]{0,0,0}\makebox(0,0)[rt]{\lineheight{0}\smash{\begin{tabular}[t]{r}node $\check{a}$\end{tabular}}}}%
    \put(0.44683926,0.36134669){\color[rgb]{1,1,1}\makebox(0,0)[t]{\lineheight{0}\smash{\begin{tabular}[t]{c}$\mu_{\check{a}}$\end{tabular}}}}%
    \put(0.51504345,0.36035877){\color[rgb]{1,1,1}\makebox(0,0)[t]{\lineheight{0}\smash{\begin{tabular}[t]{c}$\pi_{\check{a}}$\end{tabular}}}}%
    \put(0.52792335,0.44933646){\color[rgb]{0,0,0}\makebox(0,0)[t]{\lineheight{0}\smash{\begin{tabular}[t]{c}$\hat{\mu}_{\check{a}}$\end{tabular}}}}%
    \put(0.59007944,0.44933646){\color[rgb]{0,0,0}\makebox(0,0)[t]{\lineheight{0}\smash{\begin{tabular}[t]{c}$\hat{\pi}_{\check{a}}$\end{tabular}}}}%
    \put(0.59007944,0.15405661){\color[rgb]{0,0,0}\makebox(0,0)[t]{\lineheight{0}\smash{\begin{tabular}[t]{c}$\hat{\pi}_a$\end{tabular}}}}%
    \put(0.44353307,0.16719297){\color[rgb]{0,0,0}\makebox(0,0)[t]{\lineheight{0}\smash{\begin{tabular}[t]{c}$\Delta_a$\end{tabular}}}}%
    \put(0.62478799,0.03440115){\color[rgb]{0,0,0}\makebox(0,0)[rt]{\lineheight{0}\smash{\begin{tabular}[t]{r}node $a$\end{tabular}}}}%
    \put(0.62478799,0.32602758){\color[rgb]{0,0,0}\makebox(0,0)[rt]{\lineheight{0}\smash{\begin{tabular}[t]{r}node $\check{a}$\end{tabular}}}}%
    \put(0.84906584,0.36045893){\color[rgb]{0.58823529,0.58823529,0.58823529}\makebox(0,0)[t]{\lineheight{0}\smash{\begin{tabular}[t]{c}$\pi_{\check{a}}$\end{tabular}}}}%
    \put(0.86194574,0.44943674){\color[rgb]{0.58823529,0.58823529,0.58823529}\makebox(0,0)[t]{\lineheight{0}\smash{\begin{tabular}[t]{c}$\hat{\mu}_{\check{a}}$\end{tabular}}}}%
    \put(0.92410182,0.44943674){\color[rgb]{0.58823529,0.58823529,0.58823529}\makebox(0,0)[t]{\lineheight{0}\smash{\begin{tabular}[t]{c}$\hat{\pi}_{\check{a}}$\end{tabular}}}}%
    \put(0.78086173,0.06616673){\color[rgb]{0,0,0}\makebox(0,0)[t]{\lineheight{0}\smash{\begin{tabular}[t]{c}$\mu_a$\end{tabular}}}}%
    \put(0.84906584,0.06517881){\color[rgb]{0,0,0}\makebox(0,0)[t]{\lineheight{0}\smash{\begin{tabular}[t]{c}$\pi_a$\end{tabular}}}}%
    \put(0.73202709,0.20383061){\color[rgb]{1,1,1}\makebox(0,0)[t]{\lineheight{0}\smash{\begin{tabular}[t]{c}$\delta_a$\end{tabular}}}}%
    \put(0.86194574,0.15415652){\color[rgb]{0,0,0}\makebox(0,0)[t]{\lineheight{0}\smash{\begin{tabular}[t]{c}$\hat{\mu}_a$\end{tabular}}}}%
    \put(0.92410182,0.15415652){\color[rgb]{0,0,0}\makebox(0,0)[t]{\lineheight{0}\smash{\begin{tabular}[t]{c}$\hat{\pi}_a$\end{tabular}}}}%
    \put(0.77881542,0.16729288){\color[rgb]{1,1,1}\makebox(0,0)[t]{\lineheight{0}\smash{\begin{tabular}[t]{c}$\Delta_a$\end{tabular}}}}%
    \put(0.95881042,0.03450106){\color[rgb]{0,0,0}\makebox(0,0)[rt]{\lineheight{0}\smash{\begin{tabular}[t]{r}node $a$\end{tabular}}}}%
    \put(0.95881042,0.32612748){\color[rgb]{0,0,0}\makebox(0,0)[rt]{\lineheight{0}\smash{\begin{tabular}[t]{r}node $\check{a}$\end{tabular}}}}%
    \put(0.04176933,0.58194651){\color[rgb]{0,0,0}\makebox(0,0)[lt]{\lineheight{0}\smash{\begin{tabular}[t]{l}\textsf{Prediction step}\end{tabular}}}}%
    \put(0.37218822,0.58194651){\color[rgb]{0,0,0}\makebox(0,0)[lt]{\lineheight{0}\smash{\begin{tabular}[t]{l}\textsf{Update step}\end{tabular}}}}%
    \put(0.70621061,0.58194651){\color[rgb]{0,0,0}\makebox(0,0)[lt]{\lineheight{0}\smash{\begin{tabular}[t]{l}\textsf{Prediction error step}\end{tabular}}}}%
    \put(0.62305622,0.19711683){\color[rgb]{0,0,0}\makebox(0,0)[t]{\lineheight{0}\smash{\begin{tabular}[t]{c}$\Omega_a$\end{tabular}}}}%
    \put(0.44533482,0.27987174){\color[rgb]{0.91764706,0.6627451,0.70588235}\makebox(0,0)[rt]{\lineheight{0}\smash{\begin{tabular}[t]{r}$\kappa_{\check{a},a}$\end{tabular}}}}%
    \put(0.58954077,0.24199777){\color[rgb]{0,0,0}\makebox(0,0)[t]{\lineheight{0}\smash{\begin{tabular}[t]{c}$\gamma_a$\end{tabular}}}}%
    \put(0.54447937,0.29651684){\color[rgb]{0.70980392,0.76862745,0.67058824}\makebox(0,0)[lt]{\lineheight{0}\smash{\begin{tabular}[t]{l}$\kappa_{\check{a},a}^2$\end{tabular}}}}%
    \put(0.50345099,0.27987174){\color[rgb]{0.91764706,0.6627451,0.70588235}\makebox(0,0)[lt]{\lineheight{0}\smash{\begin{tabular}[t]{l}$\kappa_{\check{a},a}^2$\end{tabular}}}}%
    \put(0.13762296,0.28922109){\color[rgb]{0.3372549,0.41960784,0.5372549}\makebox(0,0)[lt]{\lineheight{0}\smash{\begin{tabular}[t]{l}$\kappa_{\check{a},a}$\end{tabular}}}}%
    \put(0.23824027,0.23069206){\color[rgb]{0.41960784,0.5372549,0.3372549}\makebox(0,0)[t]{\lineheight{0}\smash{\begin{tabular}[t]{c}$\rho_{a}$\end{tabular}}}}%
    \put(0.2212809,0.19572053){\color[rgb]{0,0,0}\makebox(0,0)[t]{\lineheight{0}\smash{\begin{tabular}[t]{c}$P_{a}$\end{tabular}}}}%
    \put(0.3061116,0.23078106){\color[rgb]{0.70980392,0.76862745,0.67058824}\makebox(0,0)[t]{\lineheight{0}\smash{\begin{tabular}[t]{c}$\omega_{a}$\end{tabular}}}}%
    \put(0.28692549,0.19577579){\color[rgb]{0,0,0}\makebox(0,0)[t]{\lineheight{0}\smash{\begin{tabular}[t]{c}$\Omega_{a}$\end{tabular}}}}%
    \put(0.15098821,0.11583901){\color[rgb]{0.3372549,0.41960784,0.5372549}\makebox(0,0)[t]{\lineheight{0}\smash{\begin{tabular}[t]{c}$\lambda_{a}$\end{tabular}}}}%
  \end{picture}%
\endgroup%

  \caption{Message-passing for volatility coupling. Interactions of two nodes, node~$a$ and its volatility parent node~$\check{a}$, are shown during the three steps of a trial (Prediction step, left; Update step, middle; Prediction error step, right). The quantities that are being computed in each step are highlighted in white. Logic of display as in figure~\ref{fig:vapeSteps}.}
  \label{\figlabel}
\end{figure}

For a node~$\check{a}$ which is the \textsf{volatility parent} of node~$a$, the update equations for computing a new posterior mean~$\mu_{\check{a}}^{(k)}$ and a new posterior precision~$\pi_{\check{a}}^{(k)}$ have been described by \citet{mathys2011}. Here, we will introduce two changes to the notation to simplify the equations themselves and their implementation:

First, we will express the volatility PE, or \textsf{VOPE}, as a function of the previously defined value PE, or \textsf{VAPE}. That means from now on, we will use the symbol $\delta$ only for \textsf{VAPE}s:
\begin{equation}
    \delta_a^{(k)} \equiv \delta_a^{(k, VAPE)} = \mu_a^{(k)} - \hat{\mu}_a^{(k)}.
\end{equation}
We use the symbol $\Delta$ for \textsf{VOPE}s, which we define as
\begin{equation}
  \begin{split}
    \Delta_a^{(k)} \equiv \delta_a^{(k, VOPE)} &:= \hat{\pi}_a^{(k)} \left( \frac{1}{\pi_{a}^{(k)}} + \left(\delta_a^{(k)}\right)^2 \right) - 1 \\
    &=  \frac{\hat{\pi}_a^{(k)}}{\pi_{a}^{(k)}} + \hat{\pi}_a^{(k)} \left(\delta_a^{(k)}\right)^2 - 1. \\
  \end{split}
\end{equation}
For a derivation of this definition based on the equations given in \citet{mathys2011}, cf.\hyperref[sec:volupdate]{Appendix~\ref*{sec:volupdate}}.\\

Second, we will introduce another quantity, which reflects the volatility-weighted precision of the prediction:
\begin{equation}
  \gamma_a^{(k)} := \Omega_a^{(k)} \hat{\pi}_a^{(k)},
\end{equation}
which will be computed as part of the \textsf{prediction step} and will be termed \textsf{effective precision of the prediction} owing to its role in the update equations. This definition serves to simplify the equations and the corresponding message passing.

With these two definitions, namely those of the \textsf{VOPE}~$\Delta^{(k)}$ and of the effective precision of the prediction~$\gamma^{(k)}$, the update equations for the precision and the mean of volatility parent~$\check{a}$ simplify to:
\vspace{0.5cm}

\noindent
\begin{minipage}{\textwidth}

\begin{align}
\pi_{\check{a}}^{\left(k\right)} &= \hat{\pi}_{\check{a}}^{\left(k\right)}
            + \frac{1}{2} \left(\kappa_{\check{a},a} \gamma_{a}^{\left(k\right)}\right)^2
            + \left(\kappa_{\check{a},a} \gamma_{a}^{\left(k\right)}\right)^2 \Delta_{a}^{\left(k\right)}
            - \frac{1}{2} \kappa_{\check{a},a}^2 \gamma_{a}^{\left(k\right)} \Delta_{a}^{\left(k\right)}\\
\mu_{\check{a}}^{\left(k\right)} &= \hat{\mu}_{\check{a}}^{\left(k\right)}
            + \frac{1}{2} \frac{\kappa_{\check{a},a} \gamma_{a}^{\left(k\right)}}{\pi_{\check{a}}^{\left(k\right)}} \Delta_{a}^{\left(k\right)}
            \label{eq:ghgfVolUpdate}
\end{align}
\vspace{1pt}

\end{minipage}
\vspace{0.5cm}

\noindent
This means that at the time of the update, volatility parent node~$\check{a}$ needs to have access to the
following quantities:
\begin{description}
\item[Its own prediction:]     $\hat{\mu}_{\check{a}}^{(k)}$, $\hat{\pi}_{\check{a}}^{(k)}$
\item[Coupling strength:]       $\kappa_{\check{a},a}$
\item[From level below:]        $\Delta_{a}^{(k)}$, $\gamma_{a}^{(k)}$
\end{description}

These equations are illustrated in the middle panel of Figure~\ref{fig:vopeSteps}. We note the structural similarities between nodes that serve as value parents and nodes that serve as volatility parents: updates of the mean are always driven by precision-weighted prediction errors, and updates of the precision require some estimate of the prediction of the precision of the child node ($\hat{\pi}_a$ or $\gamma_a$). These similarities allow us to make statements about the message passing architecture within and across nodes that generalize across coupling types (see Figure~\ref{fig:infnet}\textsf{B}).\\

An interesting difference to the implied message passing in predictive coding proposals \citep{bastos2012,shipp2016} arises from the \textsf{update step}: The HGF architecture requires that not only (precision-weighted) prediction errors are being sent bottom-up between nodes, but also estimates of prediction precision ($\hat{\pi}_a$ or $\gamma_a$) which serve to update belief precision in the higher-level node.

\subsubsection*{The prediction error step}
Finally, in the \textsf{PE step}, a node computes the deviation of its recently updated posterior from its time step-specific prediction. This can result in two different types of PEs: \textsf{VAPEs} and \textsf{VOPEs}. These will, in turn, be used to communicate with the node's parent nodes, if it has any. Therefore, this step  again depends on the nature of a node's parent nodes and can also be considered as the process of gathering all the information required
by any existing parents. In addition to the PE, parent nodes will require some estimate of the precision of the prediction (see the previous section on the \textsf{update step}). \\

If node~$a$ is the value child of node~$b$, the following quantities have to be sent up to node~$b$:
\begin{description}
\item[Precision of the prediction:]     $\hat{\pi}_{a}^{(k)}$
\item[Prediction error:]        $\delta_{a}^{(k)}$
\end{description}
Node~$a$ has already performed the \textsf{prediction step} (see above), so it  has already computed the precision of the prediction for the current time step,~$\hat{\pi}_{a}^{(k)}$. Hence, in the \textsf{PE step}, it needs to perform only the following calculation (illustrated in the right panel of Figure~\ref{fig:vapeSteps}):
\vspace{0.5cm}

\noindent
\begin{minipage}{\textwidth}

\begin{equation}
\delta_a^{(k)}  = \mu_a^{(k)} - \hat{\mu}_a^{(k)}
\end{equation}
\vspace{1pt}

\end{minipage}
\vspace{0.5cm}

\noindent

Note that $\delta_a^{(k)}$ represents a prediction error from the perspective of the parent node - the difference between the expected state of the child and the actual state at time step~$k$. From the perspective of the child node~$a$, the difference between its prior and its posterior instead represents a belief update (Bayesian surprise).  

If node~$a$ is the volatility child of node~$\check{a}$, the following quantities have to be sent up to node~$\check{a}$ (see also necessary information from level below in a volatility parent's \textsf{update step}):
\begin{description}
    \item[Effective precision of the prediction:]      $\gamma_{a}^{(k)}$
    \item[Prediction error:]        $\Delta_{a}^{(k)}$
\end{description}
Node~$a$ has already performed the \textsf{prediction step} at the previous time step, so it has already computed the precision of the prediction, $\hat{\pi}_{a}^{(k)}$, and the total predicted volatility, $\Omega_a^{(k)}$, and out of
these the effective precision of the prediction, $\gamma_{a}^{(k)}$, for the current time step. Hence, in the \textsf{PE step}, it needs to perform only the following calculations  (illustrated in the right panel of Figure~\ref{fig:vopeSteps}):
\vspace{0.5cm}

\noindent
\begin{minipage}{\textwidth}

\begin{align}
\delta_a^{(k)} &= \mu_a^{(k)} - \hat{\mu}_a^{(k)}\\
\Delta_a^{(k)} &= \frac{\hat{\pi}_a^{(k)}}{\pi_{a}^{(k)}} + \hat{\pi}_a^{(k)} \left(\delta_a^{(k)}\right)^2 - 1.
\end{align}
\vspace{1pt}

\end{minipage}
\vspace{0.5cm}

\noindent
In other words, if node~$a$ has any parents, the \textsf{VAPE} will always be computed(as it features in both scenarios), whereas the computation of a \textsf{VOPE} is only necessary if node~$a$ has a volatility parent.

Our framework allows for multiple parent nodes of either type (e.g., more than one volatility parent). However, it is important to note that all parent nodes of the same type will be sent the same bottom-up prediction errors by their child node. The only difference between the parent nodes is in their coupling strength with the child node. Thus, their relative contribution to the belief about the child node is determined by their starting value and coupling strength. This reflects the fact that the learning agent does not have direct access to these latent states and only learns about them through the information from the sensory inputs (or lower-level beliefs).

As soon as the agent has access to some other source of information about these states, they can be decoupled. For example, if two parent nodes share a child node, but one of them is additionally linked to another child node, the agent can form independent beliefs about the two parent nodes given the different bottom-up signals derived from the child nodes.

\subsection*{Differences for nonlinear value coupling}
So far, we have assumed linear value coupling in presenting the computations of value parent and children nodes. In the case of nonlinear \textsf{value coupling}, the update equations only change slightly. Specifically, in the \textsf{prediction step}, we now have the function~$g$ during the computation of the new predicted mean.
Assuming that node~$a$ is the (nonlinear) value child of nodes~$b_{1:N_{vapa}}$, the total predicted mean drift for time step~$k$ (previously equation~\ref{eq:exdrift}) will be
\begin{equation}
    P_a^{(k)} = t^{(k)} \left(\rho_a + \sum_{i=1}^{N_{vapa}} \alpha_{b_i,a}
    g_{b_i,a}\left(\mu_{b_i}^{(k-1)}\right)\right).
\end{equation}
In other words, the influence of the higher-level belief $\mu_{b_i}^{(k-1)}$ on the prediction of the lower-level belief $\hat{\mu}_a^{(k)}$ is mediated by the function $g_{b_i,a}$, just as we would expect it to be based on the generative model (equation \ref{eq:nonlin}).

In the \textsf{update step} for the value parent (previously equation~\ref{eq:linvalueupdate}), we now have:
\vspace{0.5cm}

\noindent
\begin{minipage}{\textwidth}

\begin{align}
\pi_b^{(k)} &= \hat{\pi}_b^{(k)}
            + \hat{\pi}_{a}^{(k)} \left(\alpha_{b,a}^2 g'\left(\mu_{b}^{\left(k-1\right)}\right) -
					\alpha_{b,a} g''\left(\mu_{b}^{\left(k-1\right)}\right) \delta_{a}^{\left(k\right)}\right) \\
\mu_b^{\left(k\right)} &= \hat{\mu}_b^{\left(k\right)}
            + \frac{\alpha_{b,a} g'\left(\mu_{b}^{\left(k-1\right)}\right) \hat{\pi}_{a}^{\left(k\right)}} {\pi_b^{(k)}} \delta_{a}^{(k)}
\end{align}
\vspace{1pt}

\end{minipage}
\vspace{0.5cm}

\noindent
Consequently, for the \textsf{update step}, the node~$b$ now also needs access to its own previous posterior mean $\mu_{b}^{(k-1)}$. Apart from these changes, all update equations from the previous section apply. This extension allows us to account for the fact that most states in the world interact non-linearly.\\

\subsection*{Other types of nodes}
We have thus far focused our discussion on nodes that represent beliefs about continuous states which evolve in time as random walks (whether Gaussian or auto-regressive). However, the generalized HGF can also accommodate beliefs about any type of state governed by an exponential family distribution by filtering this distribution's sufficient statistics \cite{mathys_weber_2020}. This makes it possible to track binary and categorical states, where value parents track the probabilities with which one of several possible states is occupied, or with which transitions happen. This can be applied in a range of contexts, from experimental tasks where categorical probabilities must be learned \cite{marshall_pharmacological_2016} to inversions of other classes of discrete state space generative models like the Partially Observable Markov Decision Process models often used in Active Inference \citep{parr2022}.\\

Furthermore, at the lowest level of the state hierarchy, we find states that are observable. These typically do not perform random walks, but are instead generated by their parent states independently on every time step. For continuous states in the case of value coupling, this corresponds to setting~$\lambda$ to zero so that the state only depends on its value parent and not on its own past anymore. Similarly, a volatility parent of an outcome state becomes a noise parent - because the variance in the Gaussian distribution no longer corresponds to a step size with respect to a previous mean but instead to the deviation from the current mean which is fully determined by the state's value parent(s). In other words, an agent using this type of coupling in the generative model forms an explicit and dynamic belief about the level of observation noise (stochasticity) in a particular outcome (\textsf{noise coupling}, see also Example Simulation~2, Figure~\ref{fig:example2}).

Input nodes are directly fed with observations (sensory inputs) (which can be continuous or binary) instead of receiving prediction errors from other nodes.  In many applications, we would assume that the lowest level modelled (e.g., primary sensory cortices) is already somewhat distant to the actual sensors (e.g., the retina), which means we can cast its inputs as prediction errors generated during downstream processing input (e.g., in subcortical structures). However, in the case of noise coupling (see Simulation Example~2, Figure~\ref{fig:example2}), and for the algorithmic implementation of our perceptual model, the equations goverining these nodes matter. Their treatment is presented in \hyperref[sec:input]{Appendix~\ref*{sec:input}}.

\subsection*{Summary: A network of nodes}
In this section, we have used the update equations of the \textsc{HGF} to propose a conceptualization of the inference machinery as a network of nodes which compute beliefs (i.e., probability distributions) and exchange messages with other nodes. Every node in this network represents an agent's current belief about a hidden state in its environment, on which it infers given its sensory inputs. Within every node, belief updating in response to a new input proceeds in three steps (an \textsf{update step}, a \textsf{PE step}, and a \textsf{prediction step}).

We have presented the computations for these steps for the two different kinds of coupling that the \textsc{HGF} comprises: \textsf{value coupling} and \textsf{volatility coupling}. While the update equations for volatility coupling have been derived and discussed previously \citep{mathys2011,mathys2014}, approximately Bayes-optimal inference equations for (linear and nonlinear) \textsf{value coupling} under the \textsc{HGF} have not been considered prior to our treatment here. Furthermore, our analysis identifies not only the computations entailed by each computational step, but also the message passing between nodes that is required by each step. This is interesting from a theoretical point of view, where we can compare our architecture to other proposals of belief propagation. 

From a practical point of view, the division of the belief updating machinery into subunits (nodes) allows for a modular implementation, where networks can easily be extended and modified by adding or removing nodes, or by changing the type of coupling between nodes, without having to derive the relevant equations for the whole network anew. 
In two open-source projects, we provide such an implementation \citep[in Python and Julia,][]{legrand2024,waade_in_prep_2023},
which allows users to flexibly design their own HGF structures that can be used for simulation and empirical parameter estimation. 
These tools are freely available as part of the TAPAS software collection \citep[\url{https://github.com/ComputationalPsychiatry}]{frassle_tapas_2021}.\\

Which conclusions can be drawn with respect to the message passing implied by the \textsc{hgf}?  First, while the exact computations performed during the three computational steps depend on the position of the node within the network (e.g., number of children and parent nodes) and the nature of the coupling to other nodes (\textsf{value} vs. \textsf{volatility coupling}), we have identified generic structures in these equations (see Figure~\ref{fig:infnet}\textsf{B}), which are of interest from a theoretical point of view, but also facilitate implementation. 

For example, belief updates in a node always require messages from lower-level nodes that contain prediction errors ($\delta_a$ for \textsf{value coupling}, $\Delta_a$ for volatility coupling) and estimates of precision ($\hat{\pi}_a^{(k)}$ in \textsf{value coupling} and $\gamma_{a}^{(k)}$ in volatility coupling). Similarly, forming a new prediction always entails modifying the mean of the belief by an expectation of drift, and modifying the precision by an expectation of volatility during the next time step. Expectations of drift will be driven by value parents, expectations of volatility by volatility parents, but the structure of the equations is the same for both types of coupling (see equations~\ref{eq:predict} to~\ref{eq:exvol}). 

In Figures~\ref{fig:vapeSteps} and~\ref{fig:vopeSteps}, we additionally provide a more detailed overview of what happens within and between nodes for specific coupling types. For this purpose, we additionally consider separate subunits for calculations concerning means versus precisions versus prediction errors. Exploring in how far these architectures might map onto structures and networks in biological brains will be an interesting future task.

\newpage
\section{Practical applications}
\label{sec:ghgf-examples}

In this section, we provide a few example simulations which show the range of generative models our generalization of the \textsc{hgf} encompasses. We focus on relatively low-dimensional situations that have been of interest in empirical research in cognitive neuroscience and computational psychiatry, but have so far been challenging to model. \\

  \graphicspath{{figures/fig5_example1/}}
  \def\figlabel{fig:example1}
  \begin{figure}
  \centering

  \small

  \newcommand{\w}[1]{\textcolor{white}{#1}}
  \def\svgwidth{\textwidth}

\begingroup%
  \makeatletter%
  \providecommand\color[2][]{%
    \errmessage{(Inkscape) Color is used for the text in Inkscape, but the package 'color.sty' is not loaded}%
    \renewcommand\color[2][]{}%
  }%
  \providecommand\transparent[1]{%
    \errmessage{(Inkscape) Transparency is used (non-zero) for the text in Inkscape, but the package 'transparent.sty' is not loaded}%
    \renewcommand\transparent[1]{}%
  }%
  \providecommand\rotatebox[2]{#2}%
  \newcommand*\fsize{\dimexpr\f@size pt\relax}%
  \newcommand*\lineheight[1]{\fontsize{\fsize}{#1\fsize}\selectfont}%
  \ifx\svgwidth\undefined%
    \setlength{\unitlength}{968.96818867bp}%
    \ifx\svgscale\undefined%
      \relax%
    \else%
      \setlength{\unitlength}{\unitlength * \real{\svgscale}}%
    \fi%
  \else%
    \setlength{\unitlength}{\svgwidth}%
  \fi%
  \global\let\svgwidth\undefined%
  \global\let\svgscale\undefined%
  \makeatother%
  \begin{picture}(1,0.54717608)%
    \lineheight{1}%
    \setlength\tabcolsep{0pt}%
    \put(0,0){\includegraphics[width=\unitlength]{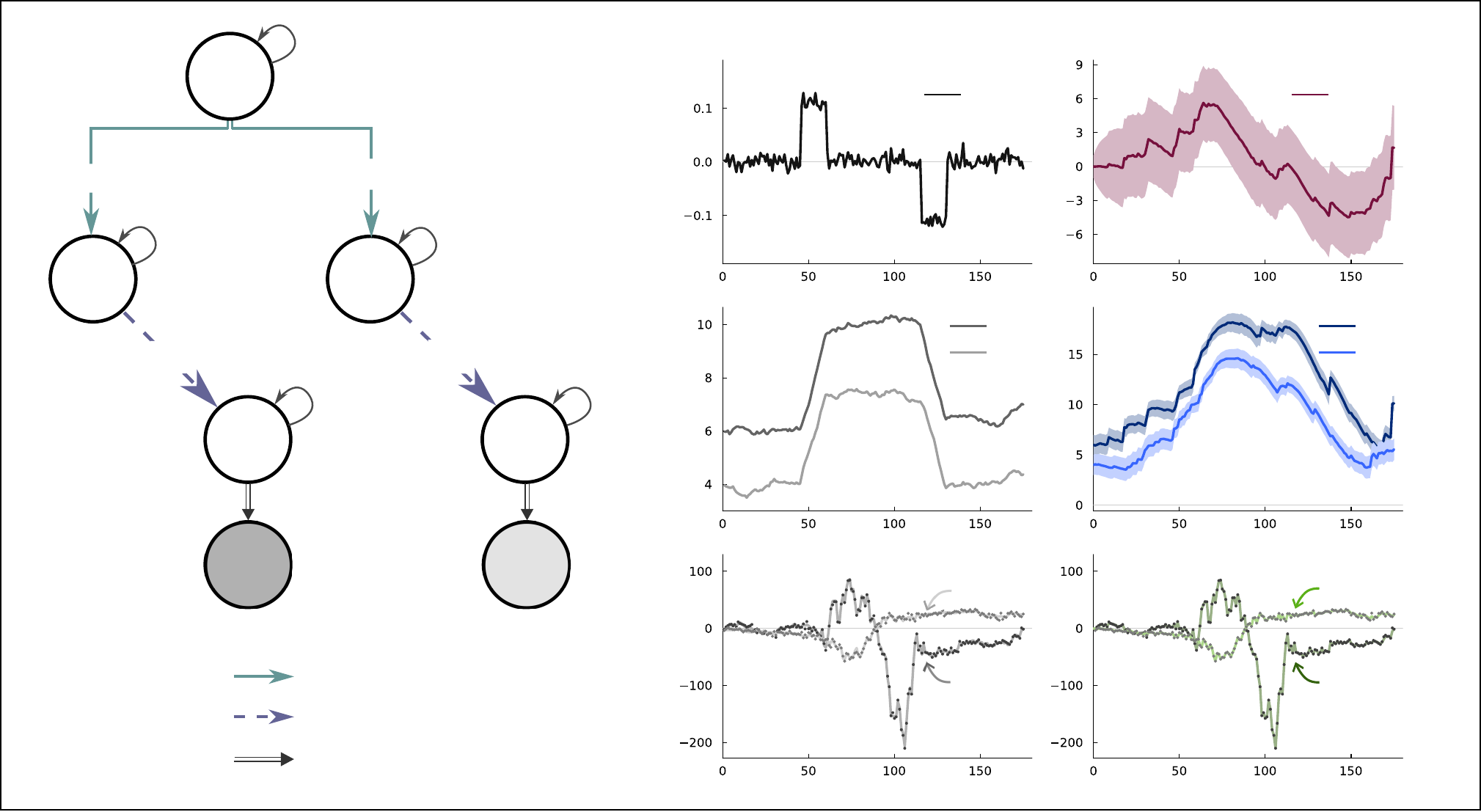}}%
    \put(0.20916865,0.08599034){\color[rgb]{0.39215686,0.58823529,0.58823529}\makebox(0,0)[lt]{\lineheight{0}\smash{\begin{tabular}[t]{l}\textsf{value coupling}\end{tabular}}}}%
    \put(0.20916865,0.05878724){\color[rgb]{0.39215686,0.39215686,0.58823529}\makebox(0,0)[lt]{\lineheight{0}\smash{\begin{tabular}[t]{l}\textsf{volatility coupling}\end{tabular}}}}%
    \put(0.20916865,0.03019025){\color[rgb]{0.19607843,0.19607843,0.19607843}\makebox(0,0)[lt]{\lineheight{0}\smash{\begin{tabular}[t]{l}\textsf{generating outcomes}\end{tabular}}}}%
    \put(0.45772467,0.52095051){\color[rgb]{0,0,0}\makebox(0,0)[t]{\lineheight{0}\smash{\begin{tabular}[t]{c}\textbf{\textsf{B}}\end{tabular}}}}%
    \put(0.71594428,0.52095051){\color[rgb]{0,0,0}\makebox(0,0)[t]{\lineheight{0}\smash{\begin{tabular}[t]{c}\textbf{\textsf{C}}\end{tabular}}}}%
    \put(0.03101416,0.52095051){\color[rgb]{0,0,0}\makebox(0,0)[t]{\lineheight{0}\smash{\begin{tabular}[t]{c}\textbf{\textsf{A}}\end{tabular}}}}%
    \put(0.1674729,0.16179681){\color[rgb]{0,0,0}\makebox(0,0)[t]{\lineheight{0}\smash{\begin{tabular}[t]{c}$u_a$\end{tabular}}}}%
    \put(0.16745183,0.24621993){\color[rgb]{0,0,0}\makebox(0,0)[t]{\lineheight{0}\smash{\begin{tabular}[t]{c}$x_a$\end{tabular}}}}%
    \put(0.35525126,0.16179681){\color[rgb]{0,0,0}\makebox(0,0)[t]{\lineheight{0}\smash{\begin{tabular}[t]{c}$u_b$\end{tabular}}}}%
    \put(0.25087325,0.35521344){\color[rgb]{0,0,0}\makebox(0,0)[t]{\lineheight{0}\smash{\begin{tabular}[t]{c}$x_{\check{b}}$\end{tabular}}}}%
    \put(0.35546865,0.24747051){\color[rgb]{0,0,0}\makebox(0,0)[t]{\lineheight{0}\smash{\begin{tabular}[t]{c}$x_b$\end{tabular}}}}%
    \put(0.06285648,0.35458263){\color[rgb]{0,0,0}\makebox(0,0)[t]{\lineheight{0}\smash{\begin{tabular}[t]{c}$x_{\check{a}}$\end{tabular}}}}%
    \put(0.25150309,0.42732941){\color[rgb]{0.39215686,0.58823529,0.58823529}\makebox(0,0)[t]{\lineheight{0}\smash{\begin{tabular}[t]{c}$\alpha_{c,\check{b}}$\end{tabular}}}}%
    \put(0.06335594,0.42578137){\color[rgb]{0.39215686,0.58823529,0.58823529}\makebox(0,0)[t]{\lineheight{0}\smash{\begin{tabular}[t]{c}$\alpha_{c,\check{a}}$\end{tabular}}}}%
    \put(0.30385261,0.30195325){\color[rgb]{0.39215686,0.39215686,0.58823529}\makebox(0,0)[t]{\lineheight{0}\smash{\begin{tabular}[t]{c}$\kappa_b$\end{tabular}}}}%
    \put(0.20151032,0.52103699){\color[rgb]{0.29411765,0.29411765,0.29411765}\makebox(0,0)[lt]{\lineheight{0}\smash{\begin{tabular}[t]{l}$\omega_c$\end{tabular}}}}%
    \put(0.21329262,0.27646247){\color[rgb]{0.29411765,0.29411765,0.29411765}\makebox(0,0)[lt]{\lineheight{0}\smash{\begin{tabular}[t]{l}$\omega_a$\end{tabular}}}}%
    \put(0.10742878,0.38482517){\color[rgb]{0.29411765,0.29411765,0.29411765}\makebox(0,0)[lt]{\lineheight{0}\smash{\begin{tabular}[t]{l}$\omega_{\check{a}}$\end{tabular}}}}%
    \put(0.40113362,0.27631921){\color[rgb]{0.29411765,0.29411765,0.29411765}\makebox(0,0)[lt]{\lineheight{0}\smash{\begin{tabular}[t]{l}$\omega_{b}$\end{tabular}}}}%
    \put(0.29685851,0.38482517){\color[rgb]{0.29411765,0.29411765,0.29411765}\makebox(0,0)[lt]{\lineheight{0}\smash{\begin{tabular}[t]{l}$\omega_{\check{b}}$\end{tabular}}}}%
    \put(0.11342171,0.30195325){\color[rgb]{0.39215686,0.39215686,0.58823529}\makebox(0,0)[t]{\lineheight{0}\smash{\begin{tabular}[t]{c}$\kappa_a$\end{tabular}}}}%
    \put(0.1552794,0.49061274){\color[rgb]{0,0,0}\makebox(0,0)[t]{\lineheight{0}\smash{\begin{tabular}[t]{c}$x_c$\end{tabular}}}}%
    \put(0.88049521,0.01463498){\color[rgb]{0.19607843,0.19607843,0.19607843}\makebox(0,0)[lt]{\lineheight{0}\smash{\begin{tabular}[t]{l}\tiny{\textsf{trial number}}\end{tabular}}}}%
    \put(0.67585835,0.1451883){\color[rgb]{0,0,0}\makebox(0,0)[t]{\lineheight{0}\smash{\begin{tabular}[t]{c}$u_b$, $x_b$\end{tabular}}}}%
    \put(0.93352692,0.1467192){\color[rgb]{0,0,0}\makebox(0,0)[t]{\lineheight{0}\smash{\begin{tabular}[t]{c}$u_b$, $\mu_b$\end{tabular}}}}%
    \put(0.66739999,0.479787){\color[rgb]{0,0,0}\makebox(0,0)[t]{\lineheight{0}\smash{\begin{tabular}[t]{c}$x_c$\end{tabular}}}}%
    \put(0.91586638,0.48026439){\color[rgb]{0,0,0}\makebox(0,0)[t]{\lineheight{0}\smash{\begin{tabular}[t]{c}$\mu_c$\end{tabular}}}}%
    \put(0.68506399,0.32408239){\color[rgb]{0,0,0}\makebox(0,0)[t]{\lineheight{0}\smash{\begin{tabular}[t]{c}$x_{\check{a}}$\end{tabular}}}}%
    \put(0.93538355,0.32408239){\color[rgb]{0,0,0}\makebox(0,0)[t]{\lineheight{0}\smash{\begin{tabular}[t]{c}$\mu_{\check{a}}$\end{tabular}}}}%
    \put(0.68495272,0.30618657){\color[rgb]{0,0,0}\makebox(0,0)[t]{\lineheight{0}\smash{\begin{tabular}[t]{c}$x_{\check{b}}$\end{tabular}}}}%
    \put(0.93527229,0.30618657){\color[rgb]{0,0,0}\makebox(0,0)[t]{\lineheight{0}\smash{\begin{tabular}[t]{c}$\mu_{\check{b}}$\end{tabular}}}}%
    \put(0.675524,0.08355717){\color[rgb]{0,0,0}\makebox(0,0)[t]{\lineheight{0}\smash{\begin{tabular}[t]{c}$u_a$, $x_a$\end{tabular}}}}%
    \put(0.9275454,0.08308202){\color[rgb]{0,0,0}\makebox(0,0)[t]{\lineheight{0}\smash{\begin{tabular}[t]{c}$u_a$, $\mu_a$\end{tabular}}}}%
    \put(0.62584292,0.01463498){\color[rgb]{0.19607843,0.19607843,0.19607843}\makebox(0,0)[lt]{\lineheight{0}\smash{\begin{tabular}[t]{l}\tiny{\textsf{trial number}}\end{tabular}}}}%
  \end{picture}%
\endgroup%

  \caption{Example simulation 1: Local versus global volatility. \textbf{A} Generative model. Global volatility state $x_c$ is a drift value parent to two local volatility states $x_{\check{a}}$ and $x_{\check{b}}$. \textbf{B} Simulated state trajectories (generative model) and observable outcomes. The two bursts in global volatility $x_c$ around trials 30 and 100 (top panel) result in an upward and downward drift in local volatilities, respectively, as seen in the middle panel. States $x_a$ and $x_b$ start off with different levels of local volatility and this difference remains throughout the simulation, demonstrating how a drift parent provides increases and decreases in the value of its children that ride on top of the child state's mean. \textbf{C} Simulated inference. Belief trajectories results from running the belief update equations on the sequence of observations $u_a$ and $u_b$. The simulated agent correctly infers on the different levels of local volatility in the two hidden states, and also detects the changes in the global volatility state (top panel). Parameters used for this simulations are given in table~\ref{tab:simus}.}
  \label{\figlabel}
\end{figure}

In the first example, we exploit the fact that nodes in the generalised HGF can share parent nodes. This allows us to model generalisation. For example, an agent who experiences high volatility in one context or domain might change their higher-level beliefs about the stability of the world more generally, and thus start to treat the dynamics in other domains or contexts as similarly volatile. In our minimal example for this, two hidden states separately generate two streams of observations (Figure~\ref{fig:example1}). These could for example represent the observed behaviours of two different acquaintances the agent interacts with, or the availability of food in two different patches. Each state also has its own phasic volatility parent. However, these volatility states are both influenced by a shared value parent that encodes a higher-level state of global environmental volatility. For example, while the availability of food varies independently over time in the two patches, the overall season might produce upticks in volatility across different food sources simultaneously. We simulate two hidden states ($x_a$ and~$x_b$), with one of them more volatile than the other (volatility parent $x_{\check{a}} > x_{\check{b}}$). The shared value (drift) parent of the two volatility nodes, $x_c$, produces a marked up-tick (fast drift starting at trial 50) and one down-tick in the individual volatilities. The simulation shows that under the chosen parameter settings the inference network is able to infer the individual as well as the shared (``global'') volatility states. The level of noise in the outcomes ($u_a$ and~$u_b$) was chosen relatively low. This model architecture could be used to explore how beliefs about environmental volatility generalize across different environmental states, and under which conditions this breaks down (e.g., overgeneralisation).\\

  \graphicspath{{figures/fig6_example2/}}
  \def\figlabel{fig:example2}
  \begin{figure}
  \centering

  \small

  \newcommand{\w}[1]{\textcolor{white}{#1}}
  \def\svgwidth{\textwidth}

\begingroup%
  \makeatletter%
  \providecommand\color[2][]{%
    \errmessage{(Inkscape) Color is used for the text in Inkscape, but the package 'color.sty' is not loaded}%
    \renewcommand\color[2][]{}%
  }%
  \providecommand\transparent[1]{%
    \errmessage{(Inkscape) Transparency is used (non-zero) for the text in Inkscape, but the package 'transparent.sty' is not loaded}%
    \renewcommand\transparent[1]{}%
  }%
  \providecommand\rotatebox[2]{#2}%
  \newcommand*\fsize{\dimexpr\f@size pt\relax}%
  \newcommand*\lineheight[1]{\fontsize{\fsize}{#1\fsize}\selectfont}%
  \ifx\svgwidth\undefined%
    \setlength{\unitlength}{1021.87064584bp}%
    \ifx\svgscale\undefined%
      \relax%
    \else%
      \setlength{\unitlength}{\unitlength * \real{\svgscale}}%
    \fi%
  \else%
    \setlength{\unitlength}{\svgwidth}%
  \fi%
  \global\let\svgwidth\undefined%
  \global\let\svgscale\undefined%
  \makeatother%
  \begin{picture}(1,0.51548316)%
    \lineheight{1}%
    \setlength\tabcolsep{0pt}%
    \put(0,0){\includegraphics[width=\unitlength]{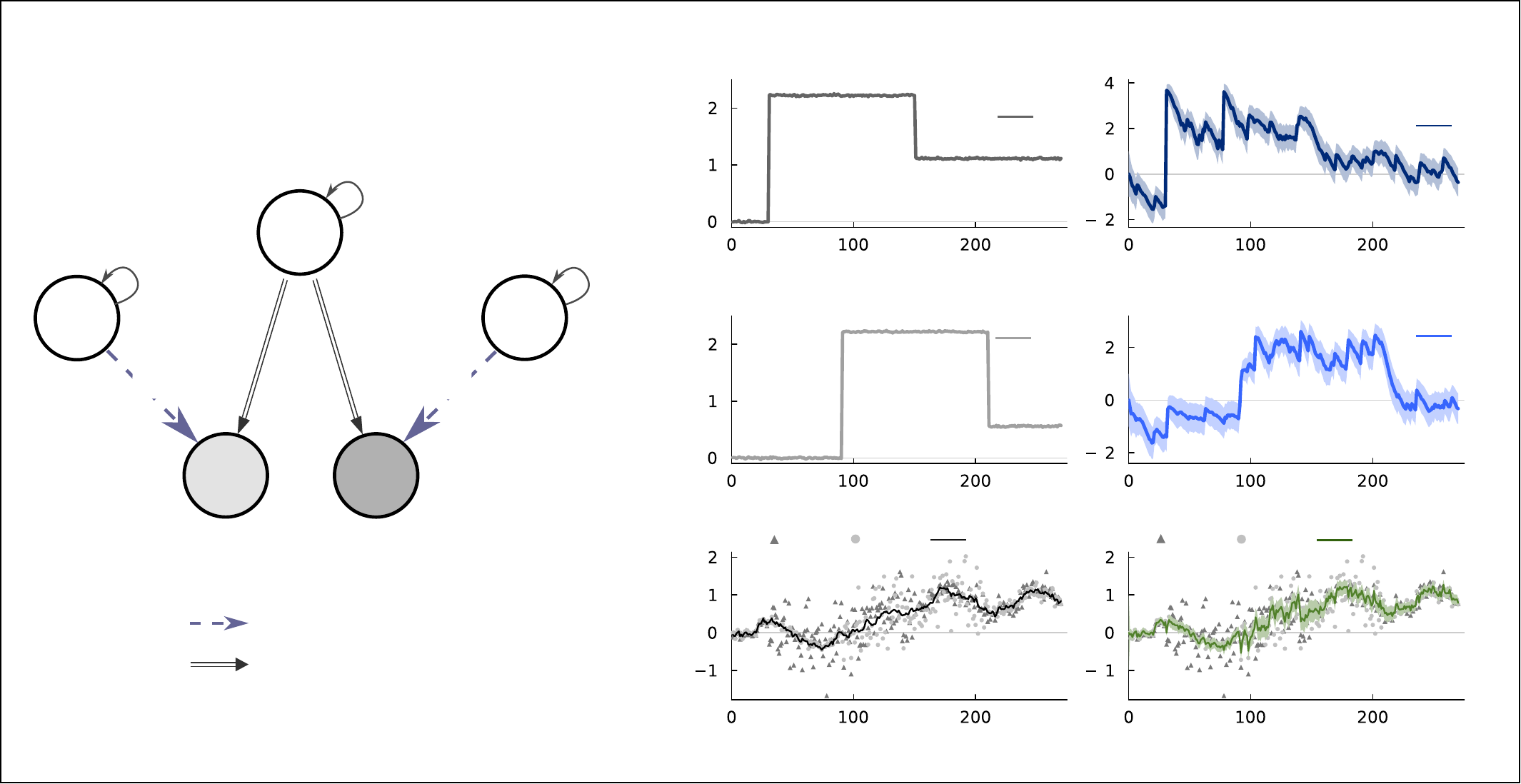}}%
    \put(0.14852753,0.19887319){\color[rgb]{0,0,0}\makebox(0,0)[t]{\lineheight{0}\smash{\begin{tabular}[t]{c}$u_a$\end{tabular}}}}%
    \put(0.24731817,0.19887319){\color[rgb]{0,0,0}\makebox(0,0)[t]{\lineheight{0}\smash{\begin{tabular}[t]{c}$u_b$\end{tabular}}}}%
    \put(0.30720476,0.25972012){\color[rgb]{0.39215686,0.39215686,0.58823529}\makebox(0,0)[t]{\lineheight{0}\smash{\begin{tabular}[t]{c}$\kappa_b$\end{tabular}}}}%
    \put(0.17353968,0.10143834){\color[rgb]{0.39215686,0.39215686,0.58823529}\makebox(0,0)[lt]{\lineheight{0}\smash{\begin{tabular}[t]{l}\textsf{noise coupling}\end{tabular}}}}%
    \put(0.19732531,0.35860357){\color[rgb]{0,0,0}\makebox(0,0)[t]{\lineheight{0}\smash{\begin{tabular}[t]{c}$x_c$\end{tabular}}}}%
    \put(0.24079291,0.38728044){\color[rgb]{0.29411765,0.29411765,0.29411765}\makebox(0,0)[lt]{\lineheight{0}\smash{\begin{tabular}[t]{l}$\omega_c$\end{tabular}}}}%
    \put(0.05063369,0.30228538){\color[rgb]{0,0,0}\makebox(0,0)[t]{\lineheight{0}\smash{\begin{tabular}[t]{c}$x_{\check{a}}$\end{tabular}}}}%
    \put(0.09289847,0.33096225){\color[rgb]{0.29411765,0.29411765,0.29411765}\makebox(0,0)[lt]{\lineheight{0}\smash{\begin{tabular}[t]{l}$\omega_{\check{a}}$\end{tabular}}}}%
    \put(0.09858115,0.25972012){\color[rgb]{0.39215686,0.39215686,0.58823529}\makebox(0,0)[t]{\lineheight{0}\smash{\begin{tabular}[t]{c}$\kappa_a$\end{tabular}}}}%
    \put(0.17353968,0.07432182){\color[rgb]{0.19607843,0.19607843,0.19607843}\makebox(0,0)[lt]{\lineheight{0}\smash{\begin{tabular}[t]{l}\textsf{generating outcomes}\end{tabular}}}}%
    \put(0.44577135,0.47778179){\color[rgb]{0,0,0}\makebox(0,0)[t]{\lineheight{0}\smash{\begin{tabular}[t]{c}\textbf{\textsf{B}}\end{tabular}}}}%
    \put(0.72732027,0.47778179){\color[rgb]{0,0,0}\makebox(0,0)[t]{\lineheight{0}\smash{\begin{tabular}[t]{c}\textbf{\textsf{C}}\end{tabular}}}}%
    \put(0.02564326,0.47778179){\color[rgb]{0,0,0}\makebox(0,0)[t]{\lineheight{0}\smash{\begin{tabular}[t]{c}\textbf{\textsf{A}}\end{tabular}}}}%
    \put(0.34591712,0.30312511){\color[rgb]{0,0,0}\makebox(0,0)[t]{\lineheight{0}\smash{\begin{tabular}[t]{c}$x_{\check{b}}$\end{tabular}}}}%
    \put(0.38952172,0.33120383){\color[rgb]{0.29411765,0.29411765,0.29411765}\makebox(0,0)[lt]{\lineheight{0}\smash{\begin{tabular}[t]{l}$\omega_{\check{b}}$\end{tabular}}}}%
    \put(0.65268088,0.15758126){\color[rgb]{0,0,0}\makebox(0,0)[t]{\lineheight{0}\smash{\begin{tabular}[t]{c}$x_c$\end{tabular}}}}%
    \put(0.7000744,0.2894939){\color[rgb]{0,0,0}\makebox(0,0)[t]{\lineheight{0}\smash{\begin{tabular}[t]{c}$x_{\check{b}}$\end{tabular}}}}%
    \put(0.69961518,0.43605578){\color[rgb]{0,0,0}\makebox(0,0)[t]{\lineheight{0}\smash{\begin{tabular}[t]{c}$x_{\check{a}}$\end{tabular}}}}%
    \put(0.58647482,0.1583271){\color[rgb]{0,0,0}\makebox(0,0)[t]{\lineheight{0}\smash{\begin{tabular}[t]{c}$u_a$\end{tabular}}}}%
    \put(0.53274153,0.15747871){\color[rgb]{0,0,0}\makebox(0,0)[t]{\lineheight{0}\smash{\begin{tabular}[t]{c}$u_b$\end{tabular}}}}%
    \put(0.8403321,0.15818786){\color[rgb]{0,0,0}\makebox(0,0)[t]{\lineheight{0}\smash{\begin{tabular}[t]{c}$u_a$\end{tabular}}}}%
    \put(0.90780214,0.15787856){\color[rgb]{0,0,0}\makebox(0,0)[t]{\lineheight{0}\smash{\begin{tabular}[t]{c}$\mu_c$\end{tabular}}}}%
    \put(0.7869961,0.15813406){\color[rgb]{0,0,0}\makebox(0,0)[t]{\lineheight{0}\smash{\begin{tabular}[t]{c}$u_b$\end{tabular}}}}%
    \put(0.97072424,0.29045917){\color[rgb]{0,0,0}\makebox(0,0)[t]{\lineheight{0}\smash{\begin{tabular}[t]{c}$\mu_{\check{b}}$\end{tabular}}}}%
    \put(0.96998045,0.4305134){\color[rgb]{0,0,0}\makebox(0,0)[t]{\lineheight{0}\smash{\begin{tabular}[t]{c}$\mu_{\check{a}}$\end{tabular}}}}%
    \put(0.88995789,0.01435262){\color[rgb]{0.19607843,0.19607843,0.19607843}\makebox(0,0)[lt]{\lineheight{0}\smash{\begin{tabular}[t]{l}\tiny{\textsf{trial number}}\end{tabular}}}}%
    \put(0.64848899,0.01435262){\color[rgb]{0.19607843,0.19607843,0.19607843}\makebox(0,0)[lt]{\lineheight{0}\smash{\begin{tabular}[t]{l}\tiny{\textsf{trial number}}\end{tabular}}}}%
  \end{picture}%
\endgroup%

  \caption{Example simulation 2: Multisensory cue combination with dynamic noise. \textbf{A}~Generative model. State $x_c$ generates two observations on each trial,~$u_a$ and~$u_b$. These could correspond to cues in different modalities, for example a visual and an auditory cue. Both observations are corrupted by noise, the level of which can change from trial to trial according to the hidden noise states~$x_{\check{a}}$and $x_{\check{b}}$. \textbf{B}~Simulated state trajectories (generative model) and observable outcomes. Both cues start off with low noise values but go on to experience periods of high and medium noise levels at different times. \textbf{C}~Simulated inference based on the sequence of observations~$u_a$ and~$u_b$. The jumps in the noise are correctly detected (upper and middle panels). When one cue becomes unreliable (e.g., between trials~25 and~90), the inference is driven relatively more by the precise cue. When both cues become noisy, the overall increase in uncertainty is reflected in the simulated agent's belief precision (lower panel, trials~90 to~150). Parameters used for this simulations are given in table~\ref{tab:simus}.}
  \label{\figlabel}
\end{figure}

In the second example, we revisit a widely studied perceptual process: multi-sensory cue combination. When humans observe two cues (e.g., an auditory and a visual one) that each provide information about a hidden state of interest~$x_c$ (such as the location of an object, Figure~\ref{fig:example2}), they can integrate both sources of information, weighing each source according to its reliability \citep{ernst2002}. Under natural circumstances, these reliabilities might dynamically change, for example due to physiological fluctuations or changes in the external environment. Here, we model a situation where two observation sequences are each subject to dynamically changing noise (hidden states~$x_{\check{a}}$ and~$x_{\check{b}}$) using nodes connected via noise coupling. Our model features a shared value parent node (hidden state~$x_c$) that generates both outcomes. The simulations show that the agent can infer these changing noise levels and adjust its inference on the shared hidden state~$x_c$ according to its current estimate of relative cue reliability. Critically, this setup allows us to model and infer changes in the subjective beliefs about the relative reliabilities and their impact on sequential belief updating. The states~$x_{\check{a}}$ and~$x_{\check{b}}$ could additionally be children of higher-level states, for example, to model higher-level beliefs about one's confidence in different sensory modalities. The model architecture is relevant for cognitive tasks that assess perceptual integration between different modalities (such as integrating interoceptive signals with auditory feedback when inferring one's own heart rate) in trial-by-trial fashion. Maladaptive over- or under-weighting of interoceptive versus exteroceptive signals is believed to drive symptoms of mental health \citep{paulus2010}. \\

Finally, in our last example, we exploit the fact that the same node in the generalised HGF can have parent or child nodes with different coupling types, and that any node within a network can be connected with an outcome node, if the state becomes somehow observable. In Figure~\ref{fig:example3}, we consider an extension of an experiment that has found widespread application in decision neuroscience: a reversal learning task, where participants track the probability of a binary outcome over a sequence of observations (e.g., whether a certain choice or stimulus is rewarded or not). The example sequence chosen here includes probability reversals and stable periods of~$p=0.5$ (i.e., both outcomes are equally likely). The typical model for this is that the participant infers on a hidden state~$x_a$ (and potentially its volatility~$x_{\check{a}}$) which represents the probability or tendency for one outcome over the other. In our example, we add an additional (continuous) observation to this hidden state~$x_a$, in other words, in addition to the binary outcomes on every trial, the agent sometimes also has access to a (noisy) sample of the continuous hidden probability state. The simulation shows that inference on small jumps in probability as well as stable periods of $50/50$ is difficult based on binary observations alone (trials~50~to~150), but that adding a continuous observation of the probability itself, even if it is only a very noisy readout of the actual probability, stabilizes inference. The model architecture thus allows us to make predictions about the relative usefulness of different types of observations when inferring on a hidden state.

  \graphicspath{{figures/fig7_example3/}}
  \def\figlabel{fig:example3}
  \begin{figure}
  \centering

  \small

  \newcommand{\w}[1]{\textcolor{white}{#1}}
  \def\svgwidth{\textwidth}

\begingroup%
  \makeatletter%
  \providecommand\color[2][]{%
    \errmessage{(Inkscape) Color is used for the text in Inkscape, but the package 'color.sty' is not loaded}%
    \renewcommand\color[2][]{}%
  }%
  \providecommand\transparent[1]{%
    \errmessage{(Inkscape) Transparency is used (non-zero) for the text in Inkscape, but the package 'transparent.sty' is not loaded}%
    \renewcommand\transparent[1]{}%
  }%
  \providecommand\rotatebox[2]{#2}%
  \newcommand*\fsize{\dimexpr\f@size pt\relax}%
  \newcommand*\lineheight[1]{\fontsize{\fsize}{#1\fsize}\selectfont}%
  \ifx\svgwidth\undefined%
    \setlength{\unitlength}{1021.87064584bp}%
    \ifx\svgscale\undefined%
      \relax%
    \else%
      \setlength{\unitlength}{\unitlength * \real{\svgscale}}%
    \fi%
  \else%
    \setlength{\unitlength}{\svgwidth}%
  \fi%
  \global\let\svgwidth\undefined%
  \global\let\svgscale\undefined%
  \makeatother%
  \begin{picture}(1,0.51548316)%
    \lineheight{1}%
    \setlength\tabcolsep{0pt}%
    \put(0,0){\includegraphics[width=\unitlength]{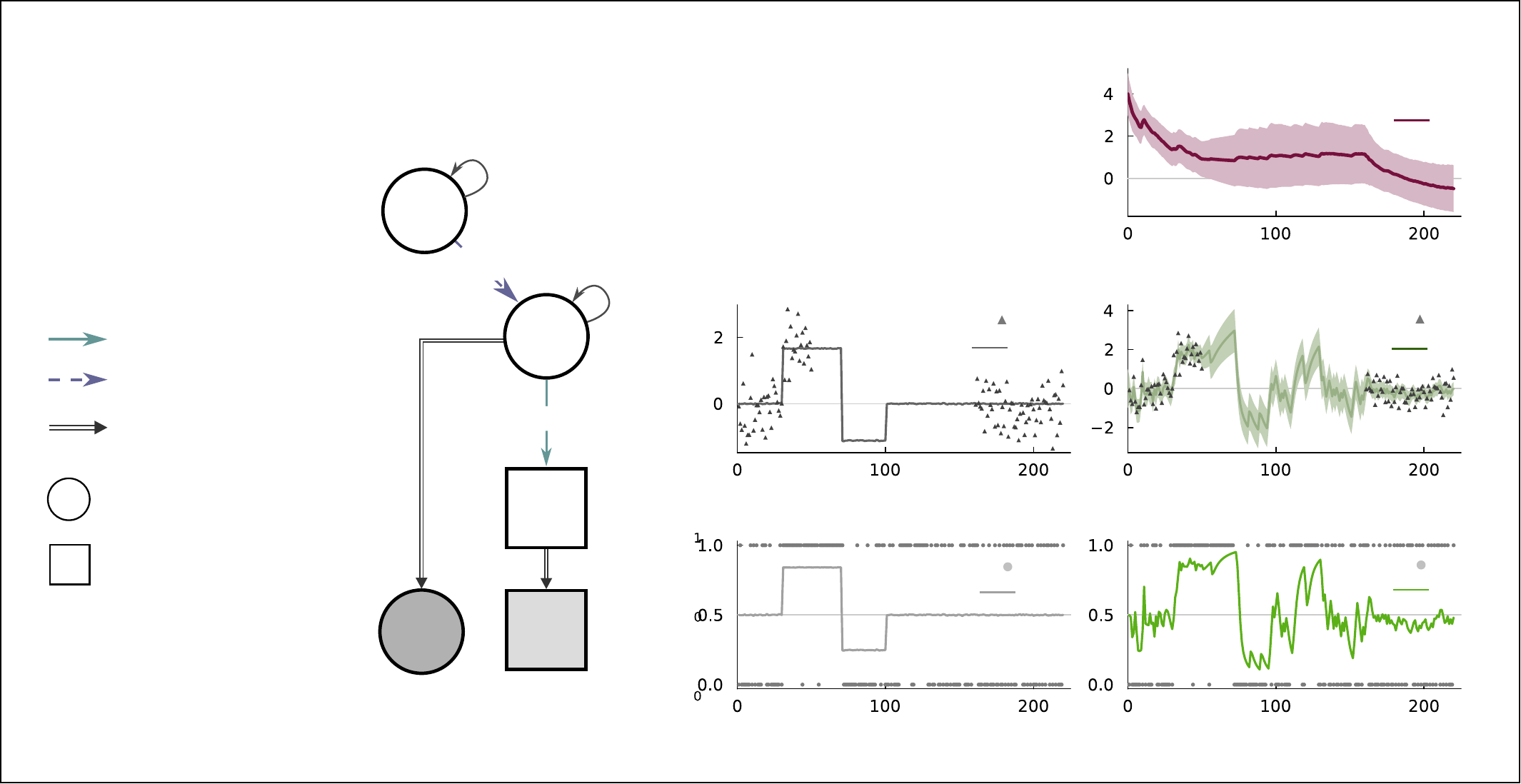}}%
    \put(0.27718468,0.09608977){\color[rgb]{0,0,0}\makebox(0,0)[t]{\lineheight{0}\smash{\begin{tabular}[t]{c}$u_a$\end{tabular}}}}%
    \put(0.08072248,0.26168166){\color[rgb]{0.39215686,0.39215686,0.58823529}\makebox(0,0)[lt]{\lineheight{0}\smash{\begin{tabular}[t]{l}\textsf{noise coupling}\end{tabular}}}}%
    \put(0.35942204,0.29030612){\color[rgb]{0,0,0}\makebox(0,0)[t]{\lineheight{0}\smash{\begin{tabular}[t]{c}$x_a$\end{tabular}}}}%
    \put(0.40288962,0.31898299){\color[rgb]{0.29411765,0.29411765,0.29411765}\makebox(0,0)[lt]{\lineheight{0}\smash{\begin{tabular}[t]{l}$\omega_a$\end{tabular}}}}%
    \put(0.27928845,0.37284979){\color[rgb]{0,0,0}\makebox(0,0)[t]{\lineheight{0}\smash{\begin{tabular}[t]{c}$x_{\check{a}}$\end{tabular}}}}%
    \put(0.32275607,0.40152666){\color[rgb]{0.29411765,0.29411765,0.29411765}\makebox(0,0)[lt]{\lineheight{0}\smash{\begin{tabular}[t]{l}$\omega_{\check{a}}$\end{tabular}}}}%
    \put(0.35936014,0.09714885){\color[rgb]{0,0,0}\makebox(0,0)[t]{\lineheight{0}\smash{\begin{tabular}[t]{c}$u_b$\end{tabular}}}}%
    \put(0.36020273,0.17633162){\color[rgb]{0,0,0}\makebox(0,0)[t]{\lineheight{0}\smash{\begin{tabular}[t]{c}$x_b$\end{tabular}}}}%
    \put(0.35933544,0.237685){\color[rgb]{0.39215686,0.58823529,0.58823529}\makebox(0,0)[t]{\lineheight{0}\smash{\begin{tabular}[t]{c}$\alpha_b$\end{tabular}}}}%
    \put(0.08072248,0.28824151){\color[rgb]{0.39215686,0.58823529,0.58823529}\makebox(0,0)[lt]{\lineheight{0}\smash{\begin{tabular}[t]{l}\textsf{value coupling}\end{tabular}}}}%
    \put(0.07967348,0.13838813){\color[rgb]{0,0,0}\makebox(0,0)[lt]{\lineheight{0}\smash{\begin{tabular}[t]{l}\textsf{binary state}\end{tabular}}}}%
    \put(0.07967348,0.18016048){\color[rgb]{0,0,0}\makebox(0,0)[lt]{\lineheight{0}\smash{\begin{tabular}[t]{l}\textsf{continuous state}\end{tabular}}}}%
    \put(0.31581059,0.33760544){\color[rgb]{0.39215686,0.39215686,0.58823529}\makebox(0,0)[t]{\lineheight{0}\smash{\begin{tabular}[t]{c}$\kappa_a$\end{tabular}}}}%
    \put(0.08072248,0.23162936){\color[rgb]{0.19607843,0.19607843,0.19607843}\makebox(0,0)[lt]{\lineheight{0}\smash{\begin{tabular}[t]{l}\textsf{generating}\end{tabular}}}}%
    \put(0.08117967,0.21142428){\color[rgb]{0.19607843,0.19607843,0.19607843}\makebox(0,0)[lt]{\lineheight{0}\smash{\begin{tabular}[t]{l}\textsf{outcomes}\end{tabular}}}}%
    \put(0.47659717,0.47778179){\color[rgb]{0,0,0}\makebox(0,0)[t]{\lineheight{0}\smash{\begin{tabular}[t]{c}\textbf{\textsf{B}}\end{tabular}}}}%
    \put(0.75814609,0.47778179){\color[rgb]{0,0,0}\makebox(0,0)[t]{\lineheight{0}\smash{\begin{tabular}[t]{c}\textbf{\textsf{C}}\end{tabular}}}}%
    \put(0.02564326,0.47778179){\color[rgb]{0,0,0}\makebox(0,0)[t]{\lineheight{0}\smash{\begin{tabular}[t]{c}\textbf{\textsf{A}}\end{tabular}}}}%
    \put(0.95623445,0.43420118){\color[rgb]{0,0,0}\makebox(0,0)[t]{\lineheight{0}\smash{\begin{tabular}[t]{c}$\mu_{\check{a}}$\end{tabular}}}}%
    \put(0.96575031,0.28394509){\color[rgb]{0,0,0}\makebox(0,0)[t]{\lineheight{0}\smash{\begin{tabular}[t]{c}$\mu_a$\end{tabular}}}}%
    \put(0.96349772,0.12503567){\color[rgb]{0,0,0}\makebox(0,0)[t]{\lineheight{0}\smash{\begin{tabular}[t]{c}$\mu_b$\end{tabular}}}}%
    \put(0.68569311,0.28402106){\color[rgb]{0,0,0}\makebox(0,0)[t]{\lineheight{0}\smash{\begin{tabular}[t]{c}$x_a$\end{tabular}}}}%
    \put(0.68534236,0.30207466){\color[rgb]{0,0,0}\makebox(0,0)[t]{\lineheight{0}\smash{\begin{tabular}[t]{c}$u_a$\end{tabular}}}}%
    \put(0.9603113,0.30259607){\color[rgb]{0,0,0}\makebox(0,0)[t]{\lineheight{0}\smash{\begin{tabular}[t]{c}$u_a$\end{tabular}}}}%
    \put(0.68739148,0.1237526){\color[rgb]{0,0,0}\makebox(0,0)[t]{\lineheight{0}\smash{\begin{tabular}[t]{c}$x_b$\end{tabular}}}}%
    \put(0.68651959,0.14040181){\color[rgb]{0,0,0}\makebox(0,0)[t]{\lineheight{0}\smash{\begin{tabular}[t]{c}$u_b$\end{tabular}}}}%
    \put(0.95843252,0.14160083){\color[rgb]{0,0,0}\makebox(0,0)[t]{\lineheight{0}\smash{\begin{tabular}[t]{c}$u_b$\end{tabular}}}}%
    \put(0.88702209,0.02903154){\color[rgb]{0.19607843,0.19607843,0.19607843}\makebox(0,0)[lt]{\lineheight{0}\smash{\begin{tabular}[t]{l}\tiny{\textsf{trial number}}\end{tabular}}}}%
    \put(0.64555319,0.02903154){\color[rgb]{0.19607843,0.19607843,0.19607843}\makebox(0,0)[lt]{\lineheight{0}\smash{\begin{tabular}[t]{l}\tiny{\textsf{trial number}}\end{tabular}}}}%
  \end{picture}%
\endgroup%

  \caption{Example simulation 3: Multimodel observations. \textbf{A}~Generative model. State~$x_a$ produces two observations on every trial: continuous observations~$u_a$ and, through binary hidden state~$x_b$, binary observations~$u_b$. This could reflect an experiment where the agent has access to binary observations, but also a continuous readout of the probability with which these observations are generated - at least on some trials. The timecourse of this probability can additionally be influenced by a volatility parent~$x_{\check{a}}$. \textbf{B}~Simulated state trajectories (generative model) and observable outcomes. The trajectory for state~$x_a$ was hand-crafted to reflect a typical experimental protocol in decision-making studies ('bandit' tasks, where state~$u_b$ corresponds to a reward outcome, and the probability of being rewarded reverses at some points during the task). Between trials~50 and~150 the participant is only presented the binary outcomes. \textbf{C}~Simulated inference. Belief trajectories result from running the belief update equations on the sequence of observations $u_a$ and $u_b$.  The simulated agent can infer on the hidden state~$x_a$, even in the absence of continuous observations, as long as the jumps/reversals are large. Picking up on the more subtle jump around trial~100 is much harder only based on binary observations. Moreover, when the true probability is around~0.5 (from trial~100 onwards), the agent tends to infer a fluctuating probability as opposed to the stable $p=0.5$, which improves once they also receive continuous observations (from trial~175), leading to a drop in estimated volatility (top panel). Parameters used for this simulations are given in table~\ref{tab:simus}.}

  \label{\figlabel}
\end{figure}

\begin{table}[]
\begin{tabular}{| c|c|c | c|c|c | c|c|c|}
\hline
\multicolumn{3}{|c|}{\textbf{Example 1}} & \multicolumn{3}{|c|}{\textbf{Example 2}} & \multicolumn{3}{|c|}{\textbf{Example 3}} \\
$\Theta$ & Process & Model      & $\Theta$ & Process & HGF       & $\Theta$ & Process & Model\\
\hline
$\omega_c$                   &\textit{n/a}&   0    & $\omega_c$                  & -3         &  -3  & $\omega_{\check{a}}$     &\textit{n/a}&  -3  \\
$\omega_{\check{a}}$         &-3          &  -3    & $\omega_{\check{a}}$        &\textit{n/a}&  -3  & $\omega_{a}$             &\textit{n/a}&  -3  \\
$\omega_{\check{b}}$         &-3          &  -3    & $\omega_{\check{b}}$        &\textit{n/a}&  -3  & $\kappa_{a}$             &  1         &   1  \\
$\omega_a$                   &-2          &  -2    & $\kappa_a $                 &  1         &   1  & $\alpha_{b} $             &  1         &   1  \\
$\omega_b$                   &-2          &  -2    & $\kappa_b $                 &  1         &   1  & $\epsilon_{a}$           &\textit{n/a}& -.5  \\
$\alpha_{c, \check{a}}$      & 2          & .05    & $\epsilon_a $               & -3         &  -3  & $x^{(0)}_{a} $           &\textit{n/a}&(0, 1)\\
$\alpha_{c, \check{b}}$      & 2          & .05    & $\epsilon_b$                & -3         &  -3  & $x^{(0)}_{\check{a}}$    &\textit{n/a}&(4, 1)\\
$\kappa_a$                   &.5          &  .5    & $x^{(0)}_c$                 &  0         &(0, 1)&                             &            &      \\
$\kappa_b$                   &.5          &  .5    & $x^{(0)}_{\check{a}}$       &\textit{n/a}&(0, 1)&                             &            &      \\
$\epsilon_a$                 & 1          &   1    & $x^{(0)}_{\check{b}}$       &\textit{n/a}&(0, 1)&                             &            &      \\
$\epsilon_b$                 & 1          &   1    &                             &            &      &                             &            &      \\
$x^{(0)}_c$                  &\textit{n/a}&(0, 1)  &                             &            &      &                             &            &      \\
$x^{(0)}_{\check{a}}$        & 6          &(6, 1)  &                             &            &      &                             &            &      \\
$x^{(0)}_{\check{b}}$        & 4          &(4, 1)  &                             &            &      &                             &            &      \\
$x^{(0)}_a$                  & 0          &(0, .5) &                             &            &      &                             &            &      \\
$x^{(0)}_b$                  & 0          &(0, .5) &                             &            &      &                             &            &  \\
\hline
\end{tabular}
\caption{Parameter values $\Theta$ used for the example simulations, for the generative process in the environment as well as the HGF's generative model. Starting states $x^{(0)}$ in the HGF's generative model are Gaussian beliefs with mean and precision ($\mu_0, \pi_0$). In all simulations, all drifts $\rho$ were set to zero, and all autoconnection strengths $\lambda$ were set to 1 (no autoregression). Values are indicated as \textit{n/a} when state trajectories have been pre-specified instead of simulated, making the parameter value irrelevant. This includes the global volatility drift $x_c$ in example 1, the two noise trajectories $x_{\check{a}}$ and $x_{\check{b}}$ in example 2, and the probability $x_{a}$ and its volatility $x_{\check{a}}$ in example 3.}
\label{tab:simus}
\end{table}
\newpage
\section{Conclusions}

The work presented here makes several contributions. First, our extension to \textsf{value coupling} includes principles of predictive coding in the HGF framework. This offers a general and versatile modelling framework, offering an approximation to optimal Bayesian inference for different types of interactions between states in the world, allowing for inter-individual differences in the dynamics of belief updating, and providing a principled treatment of the multiple forms of uncertainties agents are confronted with.

Second, we present a modular architecture for HGF networks, where beliefs represent nodes that perform three basic computational steps: an \textsf{Update} step, a \textsf{PE} step, and a \textsf{prediction} step, in response to new input. While the equations for these steps differ depending on the coupling of a node to other nodes, we identify a generic structure that allows for a modular implementation, in which nodes can easily be added to or removed from a network, without having to derive the corresponding update equations for the model anew. This feature takes significant load off researchers wanting to apply this modelling framework and to create custom models that suit their experiments. We provide such implementations in two open-source projects \citep{legrand2024,waade_in_prep_2023} which are available as part of the TAPAS software collection \citep[\url{https://github.com/ComputationalPsychiatry}][]{frassle_tapas_2021}.

Finally, by considering the case of nonlinear \textsf{value coupling} and deriving the message passing scheme implied by this, we enable a formal comparison to other proposed architectures for hierarchical Bayesian inference, most prominently Bayesian (or generalized) predictive coding \citep{bastos2012,shipp2016} and belief propagation in active inference \citep[e.g., Fig. 5.1 in][]{parr2022}. 

\subsection{Modelling different sources of uncertainty}

Whenever agents are faced with observations that violate their expectations, they need to arbitrate between different explanations -- has the world changed, requiring an update of beliefs about hidden states, or was the deviation merely due to noise in their observations? As has been shown previously \citep{mathys2011,mathys2014}, the HGF models belief updating in an agent who takes into account several forms of uncertainty for determining the optimal learning rate in the face of new observations: sensory uncertainty (how noisy are the sensory inputs I receive), informational uncertainty (how much do I already know about the hidden state that generates the inputs), and environmental uncertainty (what is the rate of change I expect in the hidden state). All of these together will determine whether (and how much) the agent updates its beliefs about a hidden state in response to unexpected observations. 

Here, we show that under the HGF, the agent cannot only learn about environmental volatility -- where higher estimates of volatility lead to faster learning, but in an undirected manner --, but also about higher-level hidden states that cause changes in lower-level hidden states in a directed fashion. For example, the weather might be more volatile in some seasons compared to others, making the agent less certain in its predictions (and faster to learn) about the likelihood of rainfall (\textsf{volatility coupling}). On the other hand, it might expect more or less rainfall in certain seasons (\textsf{value coupling}).

This flexibility in building models of hierarchically interacting states in the world allows for some  particularly interesting use cases. In Figure~\ref{fig:example1}, we have provided an example where two hidden states evolve with their own respective evolution rates (both determined by a tonic component and a phasic component), but share a higher-level value parent (``global volatility'') that drives changes in the mean of their respective phasic volatility states. This setup allows for separate estimation of "local" volatility (specific to each hidden state) and more global influences on volatility. For example, the rate of change in the availability of two different foods in the environment might vary over time and seasons in a way that is specific to each type of food, but when switching to a different environment (or after a global change to the overall climate), the availability of both foods might become more or less stable. Investigating how the brain represents each form of volatility, and thus adjusts learning rates in a modality-specific as opposed to a general manner is an important part of understanding how the brain achieves and maintains the delicate balance of precision across different hierarchical levels \citep{kanai2015,clark2013} which appears crucial for mental health \citep{petzschner2017,sterzer2018}.

Finally, agents are confronted not only with phasic variations of volatility (driving changes in the agent's environmental uncertainty), but also with dynamically changing sensory (or observation) noise. While existing modelling approaches have focused on either accounting for changes in volatility \citep{behrens2007,mathys2011,piray2020a} or block-wise changes in stochasticity or noise \citep{lee2020,nassar2010,nassar2012,nassar2016}, it has recently been pointed out that real-world agents need to able to detect (and distinguish) dynamic changes in both at the same time \citep{piray2020b}, and recent empirical work has demonstrated that human participants are indeed able to do so \citep{piray2024,foucault2025}. In the HGF, dynamic changes in observation noise can easily be accommodated by hidden states that serve as \textsf{noise parents} to observable outcome states \citep[see Appendix~\ref{sec:input} for the equations, Figure~\ref{fig:example2} for an example simulation, and][for an application to a dataset in computational psychiatry]{mikus2025}.

\subsection{Implementing the HGF's message passing scheme}

Hierarchical filtering and predictive coding are two prominent classes of hierarchical Bayesian models that cast perception as inference and model belief updates in proportion to precision-weighted prediction errors. Models from both classes are widely used, both in basic (computational) neuroscience, and for understanding mental disorders in computational psychiatry \citep{petzschner2017}.

While the message passing architecture implied by different predictive coding models has been examined in detail and partly matched with neuroanatomy and -physiology \citep[for overviews, see][]{spratling2019,keller2018,bastos2012,shipp2016}, and hierarchical filtering models share many similarities with predictive coding models, it is currently not clear whether the respective inference networks would place distinct requirements on implementation (in computers or brains), or make distinct predictions about neural readouts of perceptual inference and learning. This is partly due to their non-overlapping applications: predictive coding models consider hierarchies in which higher levels affect the mean of lower levels, and they are typically used to model inference about static sensory inputs in continuous time \citep[for a recent extension to dynamic inputs, see][]{millidge2024}.

We reduce this gap by introducing the HGF scheme for \textsf{value coupling} alongside \textsf{volatility coupling}. Our results show  \textit{(1)} that HGF inference networks for value coupling are largely compatible with recently proposed predictive coding architectures in that messages passed between nodes of the network entail a bottom-up signalling of precision-weighted prediction errors, and a top-down influence on predictions; and \textit{(2)} that there are slight but interesting differences in the updating of belief uncertainty.

One noteworthy difference between the architectures is that the update equations in the HGF require a bottom-up transmission of lower-level precision estimates (Figures~\ref{fig:infnet} and~\ref{fig:vapeSteps})\footnote{It is not surprising that the differences between the models concern the signalling of precision: The HGF derivation explicitly includes update equations for the precision associated with beliefs - as do other hierarchical Bayesian architectures based on Markovian processes \citep{Friston2013}. In contrast, most predictive coding schemes only focus on the optimization of the first moment (mode or expectation) of the posterior distribution for perceptual inference \citep{Friston2005,bogacz2017}, although the variational approach does allow approximation of the full posterior distribution including its variance.}. This is interesting, given that recent neuroanatomical studies point to additional pathways besides the classical forward (from lower-level supragranular to higher-level granular layers) and backward (from higher-level infragranular to lower-level extragranular layers) connections \citep{Markov2013,Markov2014}. Our architecture is compatible with an ascending connection within supragranular layers (for bottom-up communication of lower-level precision) that runs in parallel to a descending connection within these layers (for top-down modulation of PEs by higher-level precision), reminiscent of the ``cortical counter streams'' identified by these studies. We hope to capitalize on methodological advancements in high-resolution laminar fMRI \citep{Stephan2019,haarsma2022} in future studies to test these predictions.

\subsection{Dynamics within versus across time steps}
Although potential neurobiological implementations of approximate Bayesian inference in the brain are still hotly debated \citep{knill2004,aitchison2017}, a growing body of literature suggests that predictive coding-like architectures can account for a large range of neurophysiological findings in perception research \citep[for a recent overview, see][]{Walsh2020}, making these architectures particularly relevant for understanding human perception and inference. It would thus be interesting to examine in detail the commonalities and potential differences between message-passing schemes implied by different forms of coupling in HGF models and classical predictive coding schemes, as we have started to do here. 

Importantly, however, direct comparisons of the two models are further complicated by differences in each model's concept of time: while the HGF captures belief updates across time steps in discrete time (i.e., across sequentially arriving sensory inputs), predictive coding describes the evolution of beliefs in continuous time and, typically, in response to static sensory input \citep{Rao1999,Friston2005}\footnote{In fact, the standard form of the generative model in predictive coding does not contain temporal dynamics of the hidden states, but see \cite{friston2008,Friston2010} for extensions.}. Differential equations capture the evolution of beliefs and predictions errors and are used to simulate perceptual inference, starting with new input to the lowest level of the belief hierarchy, and ending when the ensuing PEs have been reconciled, i.e., a stable new posterior belief has emerged \citep{bogacz2017}. This can be used to make predictions about neural activity that can be compared against measurements. While, to our knowledge, existing predictive coding models have not been fit to data, the simulated neural dynamics display many features that are observed in real data, such as oscillatory tendencies, even in very small networks \citep{bogacz2017}. We refer to these simulations as \textsf{within-step dynamics} of belief updating. Furthermore, predictive coding networks have recently been rediscovered as potent candidates for solving machine learning tasks \citep[e.g., the classification or generation of static images, ][]{sun2020,song2020} and even a potential alternative to the backpropagation algorithm \citep{song2024,millidge2022predictive,whittington2017}.

On the other hand, the generative model of the HGF represents a Markovian process in discrete time; in the inference model, one-step update equations, derived based on a mean field approximation to the full Bayesian solution, quantify the change from prior to posterior on all levels of the belief hierarchy. The model is thus examined in sequential input settings to capture step-by-step learning - in other words, \textsf{across-step dynamics} of belief updating. The model provides an approximately Bayes-optimal solution to stepwise belief updating, useful for ideal observer analyses \citep{stefanics_visual_2018,weber2020,weber2022,hauke_aberrant_2022}, but the parameters of the HGF can also be fit to behavioural responses of individual participants. Model-derived agent-specific trajectories of predictions and PEs have proven particularly useful for identifying potential neural and physiological correlates of computational quantities in empirical data \citep{iglesias2013,vossel_cortical_2015,de_berker_computations_2016,diaconescu2017social,weilnhammer_neural_2018,katthagen_modeling_2018,palmer_sensorimotor_2019,deserno_volatility_2020,henco_bayesian_2020,cole_atypical_2020,lawson_computational_2021,hein_state_2021,hein_state_2022,harris_task-evoked_2022,fromm_belief_2023}. In the future, we propose to explicitly consider potential within-step dynamics of belief updates compatible with the update equations of the HGF \citep[see][for a first attempt]{weber2020diss}. Such equations could for example be derived by treating the HGF posterior values of all nodes (quantities) as the equilibrium point towards which all dynamics must converge \citep[inspired by][]{bogacz2017}. Establishing equations for within-step belief updating dynamics under the HGF might allow for empirical tests of the proposed architecture in a two-step procedure: first, individual trajectories of predictions and prediction errors are inferred from observed behaviour by fitting the HGF to participants’ responses, second, these stepwise point estimates are subsequently used to simulate expected continuous-time neuronal responses according to the differential equations.
\\

A modular implementation of the generalized HGF makes it easy to build large networks with considerable hierarchical depth, opening up exciting possibilities of applying this model architecture in machine learning applications and comparing its performance to alternative neural network architectures \citep{whittington2017,millidge2022,song2024}. When applied as a model of human cognition, however, an increase in model complexity must be matched by sufficiently rich data to successfully fit the model. In the generalised HGF, a branching out into the higher levels of the belief hierarchy (multiple parent nodes) will typically require some amount of branching out in the other direction (multiple child nodes) to disambiguate beliefs about hidden states further up.\\

In summary, we have presented a generalization of the HGF that extends its scope of hierarchical inference mechanisms to include cross-level couplings as proposed by predictive coding. The result is a new class of artificial neural networks that incorporate computational principles of two popular hierarchical Bayesian inference schemes: HGF and predictive coding. Furthermore, we have demonstrated how this extension can be cast as a modular architecture that allows for flexible changes to a model without having to re-derive update equations, thus allowing for a modular construction of complex hierarchically structured ANNs. Finally, the availability of these developments as open source software expands the toolkit of computational psychiatry and we hope that it will facilitate future investigations of perceptual inference in health and disease.

\section*{Acknowledgments}
This work was supported by the René and Susanne Braginsky Foundation (KES), the Aarhus Universitets Forskningsfond (grant AUFF-E-2019-7-10) (CM), and the Carlsberg Foundation (grant CF21-0439) (CM).


{\small
\bibliography{ghgf}
\bibliographystyle{apa-good}}


\clearpage
\section{Appendix}\label{sec:app}

\subsection{Approximate inversion for value coupling}\label{sec:appvalue}

In \cite{mathys2011}, we presented a variational approximation to the exact Bayesian inversion of our generative model which employed a mean-field approximation, and derived analytic one-step update equations using a new quadratic approximation to the variational energies. Following this procedure for the case of \textsf{value coupling}, we specify the variational energy for a value parent to derive the update equations in the main text.

The generative model for a state~$x_{a}$ with a (non)linear value parent~$x_b$ (and a volatility parent~$x_{\check{a}}$) is given by\footnote{Note that for brevity, we are omitting all priors here - strictly speaking, these equations only form a generative model if combined with appropriate priors on the model parameters and the initial states.}
\begin{equation} \label{eq:nonlin}
    x_{a}^{(k)} \sim \mathcal{N}\left(x_{a}^{(k-1)} + \alpha_{b,a} g\left(x_b^{(k)}\right), \, \exp\left(\kappa_{\check{a},a} x_{\check{a}}^{(k)} + \omega_a\right)\right),
\end{equation}
where the \textsf{value coupling} between $x_{a}$ and~$x_b$ is mediated by function~$g$, which can be nonlinear.

Using the mean-field approximation as in~\cite{mathys2011}, the variational energy and its first two derivatives for the value parent~$x_b$ are given by
\begin{equation} \label{eq:varennonlin}
\begin{split}
    I\left(x_b^{\left(k\right)}\right) =& - \frac{1}{2} \hat{\pi}_{a}^{\left(k\right)} \left(\frac{1}{\pi_{a}^{\left(k\right)}} + \left(\mu_{a}^{\left(k\right)} - \left(\mu_{a}^{\left(k-1\right)} + g\left(x_b^{\left(k\right)}\right)\right)\right)^2\right) \\
    & - \frac{1}{2} \hat{\pi}_b^{\left(k\right)} \left(x_b^{\left(k\right)} - \mu_b^{\left(k-1\right)}\right)^2 + const.\\
\end{split}
\end{equation}
\begin{equation} \label{eq:varennonlinderiv1}
\begin{split}
    I'\left(x_b^{\left(k\right)}\right) =& \hat{\pi}_{a}^{\left(k\right)} g'\left(x_b^{\left(k\right)}\right) \left(\mu_{a}^{\left(k\right)} - \left(\mu_{a}^{\left(k-1\right)} + g\left(x_b^{\left(k\right)}\right)\right)\right) \\
    & - \hat{\pi}_b^{\left(k\right)} \left(x_b^{\left(k\right)} - \mu_b^{\left(k-1\right)}\right) \\
\end{split}
\end{equation}
\begin{equation} \label{eq:varennonlinderiv2}
\begin{split}
    I''\left(x_b^{(k)}\right) &= \hat{\pi}_{a}^{(k)} \left(g''\left(x_b^{(k)}\right) \left(\mu_{a}^{(k)} - \left(\mu_{a}^{(k-1)} + g\left(x_b^{(k)}\right)\right)\right)\right) \\
    & - g'\left(x_b^{(k)}\right)^2 - \hat{\pi}_b^{\left(k\right)}\\
\end{split}
\end{equation}
We calculate the mean and precision of the Gaussian posterior for the value parent~$x_b^{(k)}$ using the rules as stated in \cite{mathys2011} (equations 38 and 40 there), which follow a quadratic approximation to the variational energy with expansion point at the posterior belief~$\mu_b^{(k-1)}$ from the previous time step. For this, we need the derivatives of the variational energies at this point:
\begin{equation} \label{eq:varennonlinderiv3}
    I'\left(\mu_b^{\left(k-1\right)}\right) = \hat{\pi}_{a}^{\left(k\right)} g'\left(\mu_b^{\left(k-1\right)}\right) \left(\mu_{a}^{\left(k\right)} - \left(\mu_{a}^{\left(k-1\right)} + g\left(\mu_b^{\left(k-1\right)}\right)\right)\right) 
\end{equation}
Here, we identify the prediction of the mean~$\hat{\mu}_{a}^{\left(k\right)}$ about the value child state~$x_{a}$ as
\begin{equation}
    \hat{\mu}_{a}^{\left(k\right)} = \mu_{a}^{\left(k-1\right)} + g\left(\mu_b^{\left(k-1\right)}\right)
\end{equation}
and thus the prediction error about $x_{a}$ as
\begin{equation}
    \delta_{a}^{\left(k\right)} = \mu_{a}^{\left(k\right)} - \left(\mu_{a}^{\left(k-1\right)} + g\left(\mu_b^{\left(k-1\right)}\right)\right).
\end{equation}
Therefore, the first derivative of the variational energy becomes
\begin{equation} \label{eq:varennonlinderiv1simpl}
    I'\left(\mu_b^{\left(k-1\right)}\right) = \hat{\pi}_{a}^{\left(k\right)} g'\left(\mu_b^{\left(k-1\right)}\right) \delta_{a}^{\left(k\right)}.
\end{equation}
Similarly, the second derivative then reads:
\begin{equation} \label{eq:varennonlinderiv2simpl}
    I''\left(\mu_b^{\left(k-1\right)}\right) = \hat{\pi}_{a}^{\left(k\right)} \left(g''\left(\mu_b^{\left(k-1\right)}\right) \delta_{a}^{\left(k\right)} - g'\left(\mu_b^{\left(k-1\right)}\right)^2\right) - \hat{\pi}_b^{\left(k\right)}
\end{equation}
With these, we can specify the update equations for the precision~$\pi_b$ and the mean~$\mu_b$ of the value parent (see \cite{mathys2011} and Appendix~B of \cite{mathys2014}):
\begin{equation} \label{eq:ghgfnonlinupdatepi}
\begin{split}
\pi_b^{\left(k\right)} &= -I''\left(\mu_b^{\left(k-1\right)}\right) \\
&= \hat{\pi}_b^{\left(k\right)} + \hat{\pi}_{a}^{\left(k\right)} \left(g'\left(\mu_b^{\left(k-1\right)}\right)^2 - g''\left(\mu_b^{\left(k-1\right)}\right) \delta_{a}^{\left(k\right)}\right)
\end{split}
\end{equation}
\begin{equation} \label{eq:ghgfnonlinupdatemu}
\begin{split}
    \mu_b^{\left(k\right)} &= \hat{\mu}_b^{\left(k\right)} + \frac{I'\left(\mu_b^{\left(k-1\right)}\right)}{\pi_b^{\left(k\right)}} \\
    &= \hat{\mu}_b^{\left(k\right)} + \frac{\hat{\pi}_{a}^{\left(k\right)} g'\left(\mu_b^{\left(k-1\right)}\right)}{\pi_b^{\left(k\right)}} \delta_{a}^{\left(k\right)} \\
\end{split}
\end{equation}
If $g(x) = x$ (linear \textsf{value coupling}), then $g'(x) = 1$ and $g''(x) = 0$, and we obtain the update equations specified in section~\ref{sec:ghgf-node}.

\subsection{Definition of a VOPE}\label{sec:volupdate}

In the main text, we introduced a new definition $\Delta$ of the volatility prediction error or \textsf{VOPE}, which we express as a function of the previously defined value prediction error $\delta$, or \textsf{VAPE}. Here, we show how our new definition derives from the definition contained in earlier work \citep{mathys2011}:

\begin{equation}
  \begin{split}
    \Delta_a^{(k)} \equiv \delta_a^{(k, VOPE)} &:= \frac{ \frac{1}{\pi_a^{(k)}} + \left(\mu_a^{(k)} - \hat{\mu}_a^{(k)}\right)^2 }{ \frac{1}{\pi_a^{(k-1)}} + \Omega_a^{(k)} } - 1 \\
    &= \hat{\pi}_a^{(k)} \left( \frac{1}{\pi_a^{(k)}} + \left(\mu_a^{(k)} - \hat{\mu}_a^{(k)}\right)^2 \right) - 1 \\
    &= \hat{\pi}_a^{(k)} \left( \frac{1}{\pi_a^{(k)}} + \left(\delta_a^{(k)}\right)^2 \right) - 1 \\
    &=  \frac{\hat{\pi}_a^{(k)}}{\pi_a^{(k)}} + \hat{\pi}_a^{(k)} \left(\delta_a^{(k)}\right)^2 - 1. \\
  \end{split} 
\end{equation}

From the first to the second line, we have used the following definition:

\begin{equation*}
\hat{\pi}_a^{\left(k\right)} := \frac{1}{ \frac{1}{\pi_a^{\left(k-1\right)}} + \Omega_a^{\left(k\right)} }.
\end{equation*}

This ensures that a given node does not need to have access to the posterior precision from the level below: $\pi_a^{(k-1)}$, which facilitates implementation.\\

In sum, we are introducing a second prediction error unit $\Delta$ which is concerned with deviations from predicted uncertainty and is informed by value prediction errors and estimates of uncertainty. It is this prediction error - a function of the squared value prediction error - which communicates between a node and its volatility parent.\\

\subsection{Computations of other types of nodes}\label{sec:input}

In the main text, we focus on continuous state nodes (i.e., states performing a Gaussian or auto-regressive random walk). We now also specify the nodalized implementation of inference for binary (and, by extension, categorical) belief nodes as well as input nodes (corresponding to observable/outcome states).

Input nodes differ from regular nodes in that the inputs do not perform a random walk ($\lambda = 0$), but are noisily generated by peripheral states at each time step (see section~\ref{sec:ghgf-mean} of the main text). We refer to the observations or sensory stimuli that enter the network as ``inputs'' and therefore call the receiving nodes input nodes. However, from the perspective of the generative model, these are the observable outputs of the network.\\

The input nodes are important elements in the \textsc{hgf} belief network. Processes which need to take place in an input node at a given time step are:
\begin{itemize}
	\item Receive a new input and store it
	\item Either receive as a second input the exact time interval since the previous input, or infer the time as 'previous plus 1' (e.g., next time step)
	\item Compute all quantities which need to be signalled to the parent node (e.g., prediction error)
	\item Send these quantities to the parent node
	\item Receive top-down messages from the parent node (e.g., $\hat{\mu}$)
	\item Compute surprise (i.e., the negative logarithm of the input's probability given the prediction)
\end{itemize}
The quantities being signalled bottom-up, and the computation of surprise, depend on the nature of the input node (continuous or binary) and on the nature of the coupling with the parent.

Because input nodes are different from HGF state nodes, but rather serve as a relay station for the input and for computing surprise, and because they capture any observation noise that might be inherent in the input, the message passing and the within-node computations differ from the generic scheme presented in the main text.

\subsubsection*{Continuous input nodes}
A continuous input node receives inputs~$u$ which can be any real number. In terms of the generative model, we think of these inputs as being sampled from a Gaussian distribution with a mean determined by the state node it is coupled to and a variance which is either constant or determined by another \textsc{hgf} node. This variance is the observation noise.

As with the coupling types introduced in the main text, we call the state node that determines the input's mean its \textsf{value parent}. However, the state node which represents the phasic component of the input's variance (or observation noise), is not a \textsf{volatility parent},  since the input has no volatility because it is not a state. Instead, we call such a parent node a \textsf{noise} parent. Every continuous input node has one \textsf{value parent}, but having a \textsf{noise parent} is optional. When noise is not determined by  a noise parent, it is a constant parameter of the input node.

\textbf{Value parents of continuous input nodes}\\
The predicted mean of an input node is simply the prediction from the value parent~$p$: 
\begin{equation}
	\hat{\mu}_u^{\left(k\right)} = \hat{\mu}_{p}^{\left(k\right)}
\end{equation}
Since the mean parent might have a drift parameter, the current prediction can only be computed once the new input has arrived. Then it needs to be signalled top-down immediately.

In the absence of a noise parent, the precision of the prediction for the input node is fully determined the input node's noise parameter $\varepsilon$:
\begin{equation}
	\hat{\pi}_u^{\left(k\right)} = \frac{1}{\exp\left(\varepsilon_u\right)}
\end{equation}
However, in the presence of a noise parent~$q$, this will additionally depend on the posterior $\mu_{q}^{(k-1)}$ of that parent at the previous time step, and the coupling parameter $\kappa_{q,u}$ of the input node with its noise parent $q$:
\begin{equation}
	\hat{\pi}_u^{\left(k\right)} = \frac{1}{\exp\left(\kappa_{q,u} \mu_{q}^{\left(k-1\right)} + \varepsilon_u\right)}.
\end{equation}

In the \textsf{update step}, the posterior mean of the input node is the input itself (the posterior precision is not required, but would be infinite, as the input is known):
\begin{align}
	\mu_u^{\left(k\right)} &= u^{\left(k\right)}
\end{align}

Finally, in the \textsf{PE step}, the value PE (or \textsf{VAPE}) will be computed as the difference the prediction and the posterior:
\begin{equation}
	\delta_u^{\left(k\right)} = \mu_u^{\left(k\right)} - \hat{\mu}_u^{\left(k\right)} = u^{\left(k\right)} - \hat{\mu}_u^{\left(k\right)}
\end{equation}
This means that prior to the update of the input node, it needs to receive the current prediction $\hat{\mu}_{p}^{(k)}$ of its parent node. \\

The update of the value parent node will look like the regular value coupling updates from previous chapters:
\begin{align}
	\pi_{p}^{\left(k\right)} &= \hat{\pi}_{p}^{\left(k\right)} + \hat{\pi}_u^{\left(k\right)}\\
	\mu_{p}^{\left(k\right)} &= \hat{\mu}_{p}^{\left(k\right)} + \frac{\hat{\pi}_u^{\left(k\right)}}{\pi_{p}^{\left(k\right)}} \delta_u^{\left(k\right)}
\end{align}
This means that the input node needs to signal bottom-up to its mean parent:
\begin{description}
	\item[Precision of the prediction:] 		$\hat{\pi}_u^{(k)}$
	\item[Prediction error:]			                $\delta_u^{(k)}$.
\end{description}
The implicit assumption here is that the connection between a continuous input node and its value parent has a connection weight of $\alpha = 1$. Should a use case arise where this is inconvenient, it can easily be changed to be a variable parameter. 

It may seem slightly artificial to construct the computational steps for the input node in this way since the actual belief about the input is represented in the parent node. However, this allows for a more modular implementation where the value parent can remain agnostic as to whether its child is another state node or a continuous input node.

Finally, to compute the surprise associated with the current input, the node needs to compute the negative logarithm of the probability of input $u^{(k)}$ under a Gaussian prediction with $\hat{\mu}_{u}^{(k)}$ as mean and $\hat{\pi}_{u}^{(k)}$ as the precision:
\begin{equation}
	-\log\left(p\left(u^{\left(k\right)}\right)\right) = \frac{1}{2} \left(\log\left(2\pi\right) - \log\left(\hat{\pi}_{u}^{\left(k\right)}\right) + \hat{\pi}_{u}^{\left(k\right)} \left(u^{\left(k\right)} - \hat{\mu}_{p}^{\left(k\right)}\right)^2\right).
\end{equation}

\textbf{Noise parents of continuous input nodes}\\
Having a noise parent for a continuous input node means that a noise PE (or \textsf{NOPE}) will be computed and signalled bottom-up during the \textsf{PE step}. We denote this PE with the symbol $\epsilon_u$. Importantly, the \textsf{NOPE} (as opposed to the \textsf{VOPE}) is not a direct function of the \textsf{VAPE}. Instead, both the posterior precision as well as the posterior mean are taken from the value parent $p$:
\begin{equation}
  \epsilon_u^{\left(k\right)} = \frac{\hat{\pi}_u^{\left(k\right)}}{\pi_{p}^{\left(k\right)}} + \hat{\pi}_u^{\left(k\right)} \left(u^{\left(k\right)} - \mu_{p}^{\left(k\right)}\right)^2 - 1. 
\end{equation}
This in turn requires that the update of the value parent happens before the computation of the \textsf{NOPE}, and the posterior of the value parent is already available to the input node.

The \textsf{update step} for the noise parent is similar to the update in volatility parents (equations~\ref{eq:ghgfVolUpdate}) with a modified prediction error and an effective precision term $\gamma_u$ fixed to 1:
\begin{align}
	\mu_{q}^{\left(k\right)} &= \hat{\mu}_{q}^{\left(k\right)} 
			+ \frac{1}{2} \frac{\kappa_{q,u} \gamma_u^{\left(k\right)}}{\pi_{q}^{\left(k\right)}} \epsilon_u^{\left(k\right)}\\
			&= \hat{\mu}_{q}^{\left(k\right)} 
			+ \frac{1}{2} \frac{\kappa_{q,u}}{\pi_{q}^{\left(k\right)}} \epsilon_u^{\left(k\right)}.
\end{align}
This similarity again means that the parent node can remain agnostic as to whether it serves as a volatility or a noise parent - as long as the input node also signals a value of $1$ as the effective precision term~$\gamma$ at every time step. Importantly, this also works for the update of the precision of the noise parent. Setting $\gamma_u$ to $1$ (and replacing the \textsf{VOPE}~$\Delta$ with the \textsf{NOPE}~$\epsilon$ and $vopa$ with $q$) in the previously established precision update for volatility parents (equation~\ref{eq:ghgfVolUpdate}) leads to:
\begin{align}
	\pi_{q}^{\left(k\right)} &= \hat{\pi}_{q}^{\left(k\right)} 
			+ \frac{1}{2} \left(\kappa_{q,u} \gamma_u^{\left(k\right)}\right)^2 
			+ \left(\kappa_{q,u} \gamma_u^{\left(k\right)}\right)^2 \epsilon_u^{\left(k\right)} 
			- \frac{1}{2} \kappa_{q,u}^2 \gamma_u^{\left(k\right)} \epsilon_u^{\left(k\right)}\\
			&= \hat{\pi}_{q}^{\left(k\right)} 
			+ \frac{1}{2} \left(\kappa_{q,u}\right)^2 
			+ \left(\kappa_{q,u}\right)^2 \epsilon_u^{\left(k\right)} 
			- \frac{1}{2} \kappa_{q,u}^2 \epsilon_u^{\left(k\right)}\\
			&= \hat{\pi}_{q}^{\left(k\right)} 
			+ \frac{1}{2} \left(\kappa_{q,u}\right)^2 
			+ \frac{1}{2} \left(\kappa_{q,u}\right)^2 \epsilon_u^{\left(k\right)}\\
			&= \hat{\pi}_{q}^{\left(k\right)} 
			+ \frac{1}{2} \left(\kappa_{q,u}\right)^2 \left(1 + \epsilon_u^{\left(k\right)}\right).
\end{align}\\

\textbf{Peculiarities of continuous input nodes and consequences for their parents}\\
The possibility for a given state node to be the value parent or the noise parent of a continuous input node has a number of consequences for the implementation of state nodes:

First, owing to the dependence of the \textsf{NOPE} on the posterior beliefs of the mean parent, the continuous input node needs to communicate with its value parent first and wait for the posteriors to be computed there and sent top-down in order to trigger an update in its noise parent.

Second, the value parent needs to send top-down not only the posterior mean, but also the posterior precision, for the same reason.

Third, the connection weight for value connections will always be $\alpha = 1$.

Fourth, for issuing a new prediction $\hat{\mu}_u$, the node needs to receive the predicted mean of its value parent at the beginning of a new time step. This means it must be possible to elicit a new prediction in regular \textsf{hgf} nodes without actually sending a prediction error, instead by only sending a new time point. The \textsf{hgf} node needs to react to this by sending top-down the new predicted mean, such that the input node can compute the PE and signal it back bottom-up for an update.\\

\noindent
Thus, the steps for a continuous input node are:
\begin{itemize}
    \item receive input $u$
    \item determine time of input
    \item send bottom-up to value parent: time of input (to elicit a prediction)
    \item receive top-down: predicted mean $\hat{\mu}_{p}$
    \item compute prediction $\hat{\mu}_u$ and retrieve $\hat{\pi}_u$
    \item compute surprise using $u$, $\hat{\mu}_u$ and $\hat{\pi}_u$
    \item compute \textsf{VAPE} using $u$ and $\hat{\mu}_u$
    \item send bottom-up to value parent: \textsf{VAPE}, $\hat{\pi}_u$, and time
    \item receive top-down: posteriors $\mu_{p}$ and $\pi_{p}$
    \item if relevant, compute \textsf{NOPE} using $u$, $\hat{\pi}_u$, $\mu_{p}$ and $\pi_{p}$
    \item send bottom-up to noise parent: \textsf{NOPE}, $\gamma_u = 1$, and time
    \item receive top-down: posterior $\mu_{q}$
    \item compute new precision of its prediction~$\hat{\pi}_u^{(k+1)}$ using its tonic observation noise $\varepsilon$ and, if present, the posterior mean of its noise parent~$\mu_{q}^{(k)}$.
\end{itemize}
\noindent
The value parent of a continuous input node needs to 
\begin{enumerate}
    \item be able to elicit new predictions based on time input
    \item send new predictions top-down immediately in the case of a continuous input node child
    \item send down not only its posterior mean, but also the precision after each update.
\end{enumerate}

\subsubsection*{Binary input nodes}
Binary input nodes serve to receive inputs that can only take on one of two values. These input nodes can only have one value parent because their stochastic properties are fully described by a Bernoulli distribution which only has one parameter. The value parents binary input nodes are binary state nodes, which are special cases of state nodes which themselves only have value parents. This implementation results in the value parents of binary \textsc{hgf} nodes being regular state nodes which can be agnostic as to whether their child node is a regular state node, a binary \textsc{hgf} node, or a continuous input node.

For binary input nodes, the observation noise is given by their noise parameter~$\varepsilon$. Therefore, the precision of the input prediction $\hat{\pi}_u$ is constant (i.e., we can treat it as a parameter). We here only present the case without observaton noise (i.e., $\varepsilon_u = 0$ or~$\hat{\pi}_u = \inf$), and leave the case of finite precision for a future treatment.\\

In general, the steps for a binary input node are:
\begin{itemize}
    \item receive input $u$
    \item determine time of input
    \item compute prediction errors, if necessary
    \item send bottom-up: $u$, input precision, and time
    \item receive top-down: prediction of parent $\hat{\mu}_{pa}$
    \item compute surprise based on message from parent.
\end{itemize}

If $\hat{\pi}_u$ is infinite, then the bottom-up messages are simply this precision $\hat{\pi}_u$ itself, and $u$. The surprise computation is also very simple:
\begin{equation}
    surprise^{\left(k\right)} = \left.
    \begin{cases}
        -\log\left(1-\hat{\mu}_{pa}^{\left(k\right)}\right), & \text{for } u^{\left(k\right)} = 1\\
        -\log\left(\hat{\mu}_{pa}^{\left(k\right)}\right), & \text{for } u^{\left(k\right)} = 0.
    \end{cases}
    \right.
\end{equation}
The special cases that follow for the update of the parent node are restricted to binary \textsc{hgf} nodes, which therefore represent their own special case of \textsc{hgf} nodes. 

\subsubsection*{Binary state nodes}
Binary nodes are parents of binary input nodes. Their cycle starts with receiving a bottom-up message from their child node, which, first of all, needs to trigger the \textsf{prediction step}. Similarly to continuous input nodes, the predictions of a binary \textsc{hgf} node depend on its parent's predictions \citep{adams2018}:
\begin{align}
    \hat{\mu}_{bin}^{\left(k\right)} &= \frac{1}{1 + \exp\left(-\kappa_{bin}\hat{\mu}_{pa}^{\left(k\right)}\right)}\\
    \hat{\pi}_{bin}^{\left(k\right)} &= \frac{1}{\hat{\mu}_{bin}^{\left(k\right)} \cdot \left(1- \hat{\mu}_{bin}^{\left(k\right)}\right)}.
\end{align}
The precision of the prediction is a direct function of the mean owing to the binary nature of the state.

Again, we need to introduce an additional top-down signalling step at the beginning of each time step, where the parent node sends down its current prediction of the mean, given the time interval since the last input.

The bottom-up message that binary state nodes receive consists of three quantities ($\hat{\pi}_u$, $u^{(k)}$, and the time of the input), and the time since the last input). The updates read:
\begin{align}
    \mu_{bin}^{\left(k\right)} &= u^{\left(k\right)}\\
    \pi_{bin}^{\left(k\right)} &= \hat{\pi}_u
\end{align}
Finally, in the \textsf{PE step}, the binary node computes a \textsf{VAPE} for its value parent:
\begin{equation}
    \delta_{bin}^{\left(k\right)} = \mu_{bin}^{\left(k\right)} - \hat{\mu}_{bin}^{\left(k\right)}.
\end{equation}
The parent node will also perform its \textsf{update step} according to the following equations \citep{mathys_uncertainty_2014,adams2018}:
\begin{equation}
    \pi_{pa}^{\left(k\right)} = \hat{\pi}_{pa}^{\left(k\right)} + \frac{\kappa^2_{bin}}{\hat{\pi}_{bin}^{\left(k\right)}}.
\end{equation}
\begin{equation}
    \mu_{pa}^{\left(k\right)} = \hat{\mu}_{pa}^{\left(k\right)} + \frac{\kappa_{bin}}{\pi_{pa}^{\left(k\right)}} \delta_{bin}^{\left(k\right)}.
\end{equation}
For the implementation, this means that we can either give the \textsc{hgf} node knowledge about who its child is and let the exact update depend on that, or we can let this be solved by the value connection, in which case this connection would need to signal the precision weight that is used for the mean update separately from the term that is used for the precision update.

In any case, the information which needs to be sent bottom-up from a binary \textsc{hgf} node to its value parent is:
\begin{description}
    \item[Prediction error:] $\delta_{bin}^{(k)}$
    \item[Predicted precision:] $\hat{\pi}_{bin}^{(k)}$
\end{description}
In case the parent does not have knowledge about its child, the information would have to be sent in the following form:
\begin{description}
    \item[Prediction error:] $\delta_{bin}^{(k)}$
    \item[Precision weight for mean update:] $1$
    \item[Precision term for precision update:] $\frac{1}{\hat{\pi}_{bin}^{(k)}}$
\end{description}

\subsubsection*{Summary: Implementational consequences}
Due to the special cases of continuous input nodes and binary state nodes, which both can be potential children of regular state nodes, we need to introduce a few changes to the update and connection logic of the regular state nodes:
\begin{itemize}
    \item  State nodes need to emit new predictions if prompted by receiving information about the time of the new input, and send this prediction top-down. This is needed both for the computation of surprise in the continuous input nodes, but also for the computation of prediction error in continuous input nodes, and for the computation of predictions in binary state nodes.
    \item In the case of value parents of continuous input nodes, state nodes need to signal top-down not only their posterior mean, but also their posterior precision.
    \item Value-coupling connections need separately to signal bottom-up the precision weight of the upcoming prediction error, and the precision term needed to update the parent's precision.
    \item Implementing noise and volatility connections in the same way allows for an implementation where regular state nodes are completely unaware about which kind of node their child is. The computation necessary for the precision update, which is more elaborate in volatility and noise coupling is then part of the connection logic.
\end{itemize}
Everything else that is unusual about the computations within binary input nodes, continuous input nodes, and binary state nodes can then be implemented within these nodes withoutaffecting the regular continuous node implementation.

\subsection{Notation overview}
 Table~\ref{table:variables} provides an overview of the notation used throughout this paper.
\begin{table}[h] 
\begin{center}
\begin{tabular}{ccc}
    Variable & Notation & Node type \\
    \hline \hline
    \multicolumn{3}{c}{\textbf{Free parameters $\Theta$}} \\
    \hline \hline
    coupling strength &  $\kappa$  &  all nodes \\
    \hline
    tonic volatility & $ \omega $  & continuous state nodes\\
    \hline
    tonic drift &  $\rho$  & \\
    \hline
    autoconnection strength &  $\lambda$  & \\
    \hline
    initial mean &  $\mu_0 $ & \\
    \hline
    initial precision &  $\pi_0$  & \\
    \hline
    tonic input noise &  $\varepsilon$  & continuous input nodes  \\
    \hline
    bias &  $b$  & \\
    
    \hline \hline
    \multicolumn{3}{c}{\textbf{Belief states $\theta$}} \\
    \hline \hline
    
    \multicolumn{3}{c}{\textit{Prediction step}} \\
    \hline

    prediction mean &  $\hat{\mu}$ & all nodes \\
    \hline
    prediction precision & $\hat{\pi}$ &  \\
    \hline
    effective precision & $\gamma$ & continuous state nodes \\
    \hline
    total predicted volatility & $\Omega$ & \\
    \hline
    total predicted drift &  $P$  &  \\
    \hline
    time since last input &  $\tau$  & \\
    \hline
    implied learning rate &  $\nu$ & categorical state nodes \\
    \hline
    predicted category probabilities &  $\xi$ & \\

    \hline
    \multicolumn{3}{c}{\textit{Belief update step}} \\
    \hline
    
    posterior mean & $\mu$ & state nodes \\
    \hline
    posterior precision &  $\pi$ &  \\
    \hline
    input value &  $u$ & input nodes \\
    
    \hline
    \multicolumn{3}{c}{\textit{Prediction error step}} \\
    \hline
    
    value prediction error &  $\delta$ & all nodes \\
    \hline
    precision prediction error &  $\Delta$ &  \\  
    \hline
    precision-weighted value prediction error &  $\psi$  &  \\
    \hline
    precision-weighted precision prediction error &  $\Psi$ & \\
    \hline
    surprise & $\Im$ &  \\
    \hline
    \\
    
\end{tabular}

    \caption{\label{table:variables} List of variables in the \textsc{hgf} and their notation. This includes the free parameters $\chi$, as well as various changing belief states.}
\end{center}
\end{table}

\end{document}